\documentclass[11pt,oneside,reqno]{amsart}

\usepackage{adjustbox, algorithm, algpseudocode, amsfonts, amssymb, amsthm, amsmath, array, authblk, bm, booktabs, braket, caption, changepage, comment, dsfont, enumitem, etoolbox, fullpage, graphicx, mathrsfs, mathtools, microtype, multicol, multirow, pifont, pythonhighlight, relsize, subfig, subfloat, textcomp, url, xpatch}

\usepackage{tikz, tikz-cd}
\usetikzlibrary{arrows.meta, positioning, shapes.geometric, decorations.pathreplacing}

\usepackage{hyperref}
\hypersetup{
	colorlinks=true,
	linkcolor=blue,
	citecolor=black,
	urlcolor=blue}
\usepackage[noabbrev,capitalise]{cleveref}
\creflabelformat{equation}{#2\textup{#1}#3}

\numberwithin{equation}{section}

\usepackage{colortbl}
\usepackage{xcolor}

\newcommand{\red}[1]{\textcolor{red}{#1}}
\definecolor{darkgreen}{rgb}{0.0, 0.75, 0.0}  
\newcommand{\darkgreen}[1]{\textcolor{darkgreen}{#1}}   

\usepackage[round]{natbib}

\setlist[enumerate]{itemsep=1ex, topsep=-1ex}
\setlist[itemize]{itemsep=0pt, topsep=0pt}

\usepackage[foot]{amsaddr}

\usepackage{titlesec}

\titleformat{\section}
    {\titlerule\vspace{-2ex}}
    {\thesection.}{1ex}{\centering\scshape}
    [\vspace{.5ex}\titlerule]

\titleformat{\subsection}[runin]
  {\normalfont\normalsize\bfseries}{\thesubsection}{1ex}{\protect\subsectiontitle}

\titleformat{\subsubsection}[runin]
  {\normalfont\normalsize\itshape}{\thesubsubsection}{1ex}{\protect\subsubsectiontitle}

\titlespacing*{\subsection}{0pt}{0.0\baselineskip}{0.5ex}
\titlespacing*{\subsubsection}{0pt}{0.0\baselineskip}{0.5ex}

\newcommand{\subsectiontitle}[1]{#1}
\newcommand{\subsubsectiontitle}[1]{#1}

\newtheorem{theorem}{Theorem}[section]

\newtheoremstyle{style}
  {\baselineskip} 
  {0em} 
  {\itshape} 
  {} 
  {\bfseries} 
  {.} 
  {.5em} 
  {} 
\theoremstyle{style}

\newtheorem*{theorem*}{Theorem}

\numberwithin{equation}{section}

\newtheorem*{definition*}{Definition}

\newtheorem*{lemma*}{Lemma}

\newtheorem*{prop*}{Proposition}

\crefname{condition}{Condition}{Conditions}

\algrenewcommand\algorithmicrequire{\textbf{Input:}}
\algrenewcommand\algorithmicensure{\textbf{Output:}}

\DeclareMathOperator{\1}{\mathds{1}}

\DeclareMathOperator*{\argmin}{arg\,min}
\newcommand{\efp}{\mathtt{efp}}
\newcommand{\nhalf}{\lfloor n/2 \rfloor}

\newcommand{\lmax}{\lambda_{\mathrm{max}}}
\newcommand{\lmin}{\lambda_{\mathrm{min}}}

\usepackage{ifthen}
\newboolean{showfigures}
\setboolean{showfigures}{true}

\title{{\Large{\LARGE N}onparametric IPSS} \\
\vspace*{0.5em}
{\small Fast, flexible feature selection with false discovery control}}

\author{
  {\large O}mar {\large M}elikechi\textsuperscript{1},
  {\large D}avid {\large B}. {\large D}unson\textsuperscript{2},
  and {\large J}effrey {\large W}. {\large M}iller\textsuperscript{1}
}
\address{\textsuperscript{1}Department of Biostatistics, Harvard T.H. Chan School of Public Health, Boston, MA}
\address{\textsuperscript{2}Department of Statistical Science, Duke University, Durham, NC}
\email{omar.melikechi@gmail.com}

\begin{document}

\frenchspacing

\maketitle

\textbf{Publication notice.} This version of the paper has been peer-reviewed and published in \textit{Bioinformatics}. See \url{https://doi.org/10.1093/bioinformatics/btaf299
}.

\begin{abstract}
Feature selection is a critical task in machine learning and statistics. However, existing feature selection methods either (i) rely on parametric methods such as linear or generalized linear models, (ii) lack theoretical false discovery control, or (iii) identify few true positives. Here, we introduce a general feature selection method with finite-sample false discovery control based on applying integrated path stability selection (IPSS) to arbitrary feature importance scores. The method is nonparametric whenever the importance scores are nonparametric, and it estimates $q$-values, which are better suited to high-dimensional data than $p$-values. We focus on two special cases using importance scores from gradient boosting (\texttt{IPSSGB}) and random forests (\texttt{IPSSRF}). Extensive nonlinear simulations with RNA sequencing data show that both methods accurately control the false discovery rate and detect more true positives than existing methods. Both methods are also efficient, running in under 20 seconds when there are 500 samples and 5000 features. We apply \texttt{IPSSGB} and \texttt{IPSSRF} to detect microRNAs and genes related to cancer, finding that they yield better predictions with fewer features than existing approaches.
\end{abstract}


\vspace{1em}
\section{Introduction}\label{sec:intro}


Identifying the important features in a dataset can greatly improve performance and interpretability in machine learning and statistical problems \citep{survey}. For example, in genomics, often only a small fraction of genes (features) are related to a disease of interest (response). By identifying these genes, scientists can save time and resources while gaining insights that would be difficult to uncover otherwise \citep{survey}. 

The goal of feature selection is to maximize the number of important features selected (true positives)---informally referred to here as \textit{power}---while minimizing the number of unimportant features selected (false positives). In simulations, we find that popular methods without theoretical false discovery control, such as recursive feature elimination and Boruta \citep{boruta}, often select many false positives (\cref{sec:simulation_studies}). 
Meanwhile, popular methods with false discovery control, namely stability selection \citep{mb,shah} and model-X knockoffs \citep{knockoffs}, have low power in simulations and select few features in practice (\cref{sec:simulation_studies,sec:cancer}).

One reason for stability selection's low power is its relatively weak theoretical upper bounds on the expected number of false positives, E(FP). Recently, \citet{ipss} proved that much stronger bounds hold for \textit{integrated path stability selection} (IPSS), which consequently identifies more true positives. Until now, IPSS has only been applied to generalized linear models, limiting its applicability since parametric assumptions are often violated or difficult to verify in practice.

Thus, there is a need for nonparametric feature selection methods with theoretical false discovery control and greater power. In this work, we address this need by applying IPSS to arbitrary feature importance scores. The result is a general feature selection method with tight upper bounds on E(FP) that are characterized by novel quantities called \textit{efp scores}. In addition to controlling E(FP), efp scores approximately control the false discovery rate (FDR) and estimate $q$-values for each feature, which are more reliable than $p$-values in genomics and other high-dimensional settings \citep{storey_gwas}.

Our proposed method is nonparametric whenever the feature importance scores come from nonparametric models. We develop two specific instances of this: IPSS for gradient boosting (\texttt{IPSSGB}) and IPSS for random forests (\texttt{IPSSRF}). Like knockoffs, neither \texttt{IPSSGB} nor \texttt{IPSSRF} assume a specific functional relationship between the response and the features. Unlike knockoffs, neither method requires knowledge of the joint distribution of the features.

In simulations, we find that \texttt{IPSSGB} and \texttt{IPSSRF} provide a better balance of FDR control and power than 12 other methods, and that \texttt{IPSSGB} performs best overall. In particular, both methods significantly outperform parametric versions of IPSS when the parametric assumptions are violated. In \cref{sec:cancer}, we find that \texttt{IPSSGB} and \texttt{IPSSRF} successfully identify microRNAs and genes related to ovarian cancer and glioma, achieving better predictive performance than other feature selection methods while using fewer features. Both are also computationally efficient.

Finally, another important aspect of feature selection is stability---the consistent selection of features across similar settings. Stability improves reproducibility, which is critical in many applications \citep{robust_rfe}. As their names suggest, stability selection and IPSS are designed to produce stable results by repeatedly applying a baseline feature selection algorithm to random subsamples of the data (\cref{sec:ipss}). \cite{nogueira} show that stability selection can produce significantly more stable results than its baseline algorithm. In this work, we focus on false discovery control and power; further study of the stability of IPSS, stability selection, and other stability-inspired methods like StabML-RFE \citep{robust_rfe} is left to future work.

\textit{Organization}. In \cref{sec:methods}, we introduce efp scores, review IPSS, and present its extension to feature importance scores. In \cref{sec:simulation_studies}, we present our simulation studies, and in \cref{sec:cancer}, we analyze ovarian cancer and glioma data. We conclude in \cref{sec:discussion} with a discussion.


\section{Methods}\label{sec:methods}


In \cref{sec:efp_scores}, we introduce efp scores. These quantities, assigned to each feature in the dataset, are used to perform feature selection with E(FP) control. In \cref{sec:ipss}, we introduce IPSS, which is a general approach for constructing efp scores. In \cref{sec:ipss_fis}, we describe how IPSS can be applied to any feature importance score, focusing in particular on importance scores from tree-based methods such as gradient boosting and random forests.

\textit{Notation}. Throughout this work, $n$ and $p$ are the number of samples and features, respectively, and $Z_{1:n}=(Z_1,\ldots,Z_n)$ is a collection of independent and identically distributed (iid) random vectors $Z_i = (X_i,Y_i)$, where each $X_i \in \mathbb{R}^p$ is a vector of features and $Y_i\in\mathbb{R}$ is a response variable. Features are identified by their indices $j\in\{1,\dots,p\}$, $\1$ is the indicator function, that is, $\1(A)=1$ if $A$ is true and $\1(A)=0$ otherwise, $\lfloor\cdot\rfloor$ is the floor function, and E and $\mathbb{P}$ are expectation and probability, respectively. The iid assumption is standard in the feature selection literature, and is assumed by stability selection, knockoffs, and IPSS. It is also a reasonable assumption in many applications involving tabular data, which is our focus in this work. For example, genomic data from unrelated individuals are widely treated as independent.


\subsection{efp scores.}\label{sec:efp_scores}


Suppose $S\subseteq \{1,\ldots,p\}$ is an unknown subset of important features that we wish to estimate using $Z_{1:n}$. An \textit{efp (expected false positive) score} is a function $\efp_{Z_{1:n}}:\{1,\dots,p\}\to[0,\infty)$ that depends on $Z_{1:n}$ and satisfies the following:
\begin{align*}
	\text{For all $t\geq 0$, if $\hat{S}(t) = \{j : \efp_{Z_{1:n}}(j) \leq t\}$ then $\mathrm{E}(\mathrm{FP}(t)) \leq t$,}
\end{align*}
where $\mathrm{E}(\mathrm{FP}(t)) = \mathrm{E}\lvert \hat{S}(t) \cap S^c\rvert$ is the expected number of false positives in $\hat{S}(t)$. That is, the estimator $\hat{S}(t)$ of $S$ selects at most $t$ false positives on average. A trivial example of an efp score is $\efp_{Z_{1:n}}(j)=p$ for all $j$. This corresponds to selecting either no features or all features. Specifically, if $t\in[0, p)$, then $\hat{S}(t) = \varnothing$ and $\mathrm{E}(\mathrm{FP}(t)) = 0 \leq t$, while if $t\in [p,\infty)$, then $\hat{S}(t) = \{1,\dots,p\}$ and $\mathrm{E}(\mathrm{FP}(t)) \leq t$, since the number of false positives is at most $p$. 

The quality of an efp score is measured by the tightness of its bounds $\mathrm{E}(\mathrm{FP}(t))\leq t$. Better efp scores have tighter bounds because tight bounds enable accurate false positive control via the parameter $t$. Accurate control in turn leads to more true positives in $\hat{S}(t)$ since weak bounds overestimate the number of false positives, reducing the total number of features selected.

E(FP) and efp scores are related to two other quantities of significant interest: the false discovery rate (FDR) and $q$-values. Informally, the \textit{false discovery rate} is the expected ratio between the number of false positives and the total number of features selected, $\mathrm{FDR} = \mathrm{E(FP/(TP + FP))}$, and the \textit{$q$-value of feature $j$} is the smallest FDR when $j$ is selected \citep{storey_pfdr}. When $p$ is large, as is often the case in genomics, we have
\begin{align}\label{eq:pfdr}
\begin{split}
	\mathrm{pFDR}(t) &\approx \mathrm{FDR}(t)
		\approx \frac{\mathrm{E}(\mathrm{FP}(t))}{\mathrm{E}\lvert\hat{S}(t)\rvert}
		\leq \frac{t}{\mathrm{E}\lvert\hat{S}(t)\rvert},
\end{split}
\end{align}
where $\mathrm{pFDR}(t)=\mathrm{E}(\mathrm{FP}(t) / \lvert\hat{S}(t)\rvert \mid \lvert\hat{S}(t)\rvert > 0)$ is the \textit{positive false discovery rate}, the two approximations are from \citet{storey_pfdr}, and the inequality holds by the definition of an efp score. It follows that the $q$-value of feature $j$ satisfies
\begin{align}\label{eq:qvalue}
\begin{split}
	q_j &= \inf_{\{t : \efp(j) \leq t\}} \mathrm{pFDR}(t)
		\lesssim \inf_{\{t : \efp(j) \leq t\}} \frac{t}{\mathrm{E}\lvert\hat{S}(t)\rvert},
\end{split}
\end{align}
where the equality is the definition of the $q$-value (here, $\efp(j)$ denotes the observed value of the test statistic $\efp_{Z_{1:n}}(j)$) \citep{storey_pfdr}, and the approximate inequality $\lesssim$ holds by \cref{eq:pfdr}. Thus, when the efp score has tight bounds, the $q$-value of feature $j$ is well-approximated by the rightmost term in \cref{eq:qvalue}, which is easily estimated in practice by replacing $\mathrm{E}\lvert\hat{S}(t)\rvert$ with $\lvert\hat{S}(t)\rvert$. Similarly, by \cref{eq:pfdr}, $\mathrm{FDR}(t)$ is approximately bounded by $t/\lvert\hat{S}(t)\rvert$. So, as an alternative to specifying the target E(FP) parameter $t$, one can control the FDR at level $\alpha$ by choosing the largest set $\hat{S}(t)$ such that $t/\lvert\hat{S}(t)\rvert \leq \alpha$. The largest such set is chosen to maximize true positives.


\subsection{Integrated path stability selection.}\label{sec:ipss}


Integrated path stability selection (IPSS) constructs efp scores by applying baseline feature selection algorithms to random subsamples of the data. Specifically, let $S$ be an unknown subset of true features as before, and let $\hat{S}_\lambda\subseteq\{1,\ldots,p\}$ be an estimator of $S$ that depends on the data and a parameter $\lambda > 0$. Note that $\hat{S}_\lambda$ and $\hat{S}(t)$ are distinct estimators of $S$: The former is a baseline algorithm whose parameter $\lambda$ appears as a subscript.

The IPSS subsampling procedure, which is also used in stability selection, consists of $B$ subsampling iterations. Each iteration consists of randomly drawing disjoint subsets $A_1,A_2\subseteq\{1,\dots,n\}$ of size $\nhalf$ and evaluating $\hat{S}_\lambda(Z_{A_1})$ and $\hat{S}_\lambda(Z_{A_2})$ at all $\lambda$ in some interval $\Lambda\subseteq (0,\infty)$, where $Z_A = (Z_i:i\in A)$ (the choice of $\nhalf$ samples is needed for existing stability selection theorems---not just IPSS---to hold). After $B$ iterations, the \textit{estimated selection probability} $\hat\pi_j(\lambda) = \frac{1}{2 B}\sum_{b=1}^{2B} \1(j\in\hat{S}_\lambda(Z_{A_b}))$ of feature $j$ is the proportion of times $j$ is selected over all $2B$ subsets. Large values of $\hat{\pi}_j(\lambda)$ correspond to $j$ being selected by $\hat{S}_\lambda$ on many of the random subsamples, suggesting that $j$ is important.

\cite{ipss} prove that for any $\Lambda\subseteq (0,\infty)$, any probability measure $\mu$ on $\Lambda$, and certain functions $f:[0,1]\to\mathbb{R}$, the function $\efp_{Z_{1:n}}:\{1,\dots,p\}\to[0,p]$ defined by
\begin{align}\label{eq:efp}
	\mathtt{efp}_{Z_{1:n}}(j) &= \min\left\{\frac{\mathcal{I}(\Lambda)}{\int_\Lambda f(\hat{\pi}_j(\lambda))\mu(d\lambda)},\ p\right\},
\end{align}
is a valid efp score, that is, $\hat{S}(t) = \{j : \efp_{Z_{1:n}}(j) \leq t\}$ satisfies $\mathrm{E}\lvert \hat{S}(t)\cap S^c\rvert \leq t$. An explicit form of $\mathcal{I}(\Lambda)$, which comes from the theory of IPSS, is provided in \cref{sup_sec:ipss}.

Several choices of $f$ in \cref{eq:efp} yield provably valid efp scores (see \cref{sup_sec:ipss} as well as Theorems 4.1, 4.2, S3.2, S3.3 in \cite{ipss} for detailed theoretical results about IPSS with different functions). We always use $f(x)=(2x-1)^3\1(x\geq 0.5)$ because the resulting bounds $\mathrm{E}(\mathrm{FP}(t)) \leq t$ are the tightest of any existing version of stability selection \citep{ipss}. In addition to its theoretical justification, the empirical results in \cref{fig:sensitivity_f_reg,fig:sensitivity_f_class} show that this choice of $f$ produces the best results among several functions for which valid efp scores are available. Further details about IPSS, including descriptions of $\Lambda$, $\mu$, and a derivation of \cref{eq:efp}, are provided in \cref{sup_sec:ipss}.


\subsection{IPSS for feature importance scores.}\label{sec:ipss_fis}


Until now, IPSS has only been applied to parametric baseline estimators $\hat{S}_\lambda$. A canonical example is $\ell^1$-regularized estimators $\hat{S}_\lambda = \{j : \hat{\beta}_j(\lambda) \neq 0\}$, where
\begin{align}\label{eq:l1}
	\hat\beta(\lambda) &= \argmin_{\beta\in\mathbb{R}^p} \sum_{i=1}^n \mathcal{L}(Y_i,\beta^\mathtt{T}X_i) + \lambda\sum_{j=1}^p\lvert\beta_j\rvert.	
\end{align}
Here, $\lambda > 0$ controls the strength of the $\ell^1$-penalty and $\mathcal{L}$ is a log-likelihood that assumes a specific relationship between the features and response. By extending IPSS to feature importance scores, we no longer require such restrictive assumptions.

An \textit{importance function} is a (possibly random) map $\Phi_{Z_{1:n}}:\{1,\dots,p\}\to\mathbb{R}$ that uses the data to assign an \textit{importance score} $\Phi_{Z_{1:n}}(j)$ to each feature. The possible randomness, which is additional to the randomness in $Z_{1:n}$, can come from, for example, random subsampling in tree-based algorithms. Suppressing $Z_{1:n}$ from the notation for now, assume $\Phi(j) < \Phi(k)$ means that $j$ is less important than $k$ according to $\Phi$. For example, in linear regression where $Z_i = (X_i,Y_i)$ with $Y_i = \beta^\mathtt{T}X_i + \epsilon_i$, a viable importance function is the magnitude of the estimated regression coefficient, $\Phi(j)=\lvert\hat\beta_j\rvert$. 

Associated to any importance function $\Phi$ is a baseline feature selection estimator $\hat{S}_\lambda = \{j : \Phi(j) \geq \lambda\}$, obtained by simply thresholding the importance scores. That is, given $\Phi$ and a threshold $\lambda$, $\hat{S}_\lambda$ selects the features whose importance scores are at least $\lambda$. With $\hat{S}_\lambda$ defined, we have all that is needed to implement IPSS. Furthermore, all of the theoretical results from \cite{ipss} apply without modification.


\subsubsection{Implementation.}\label{sec:implementation}


\cref{alg:ipss} outlines IPSS for feature importances. The main steps are as follows: First, features are preselected according to the procedure described in \cref{sup_sec:preselection}. This is a common preliminary step in many feature selection algorithms. Next, importance scores are evaluated for the preselected features using random, disjoint halves of the data. This process is repeated $B$ times, yielding $2B$ importance scores for each feature. We then construct a grid of $\lambda$ values. The largest, $\lmax$, is the maximum importance score across all features and all $2B$ sets of scores (hence, $\hat{S}_{\lmax} = \varnothing$). Next, starting from $\lmin=\lmax$, decrease $\lmin$ one step at a time, usually on a log scale, iteratively updating the integral $\mathcal{I}([\lmin,\lmax])$ at each step in the form of a Riemann sum approximation until $\mathcal{I}([\lmin,\lmax])$ surpasses a cutoff $C$. Once $C$ is surpassed, the $\mathtt{while}$ loop stops and the feature-specific selection probabilities and integrals are evaluated. The algorithm outputs efp scores for each feature, which are used to control E(FP), FDR, and compute $q$-values, as described in \cref{sec:efp_scores}.

Sensitivity analyses in \cref{sup_sec:sensitivity} show that IPSS depends little on $C$ and the number of subsamples $B$ (in related work, \cite{shah} also report that stability selection is insensitive to $B$). Our defaults are $C=0.05$ and $B=100$. For $\mu$, we consider the family of measures $\mu_\delta(d\lambda) = z^{-1}_\delta\lambda^{-\delta}d\lambda$, where $\delta\in\mathbb{R}$ and $z_\delta = \int_\Lambda \lambda^{-\delta}d\lambda$ is a normalizing constant that is easily computed in closed form \citep{ipss}. The values $\delta=0$ and $\delta=1$ correspond to averaging over $\Lambda$ on linear and log scales, respectively. Like $C$ and $B$, sensitivity analyses in \cref{sup_sec:sensitivity} show that IPSS is robust to the choice of $\delta$. In regression problems, we use $\delta=1.25$ for \texttt{IPSSGB} and \texttt{IPSSRF}. In classification, we use $\delta=1$ for \texttt{IPSSGB} and $\delta=1.25$ for \texttt{IPSSRF}.


\subsubsection{\texttt{IPSSGB} and \texttt{IPSSRF}.}\label{sec:ippsgb_ipssrf}


Any importance function can be combined with IPSS, providing many possible directions for future work. In this paper, we focus on importance functions from tree ensemble methods because they are nonparametric, computationally efficient, and produce state-of-the-art results on tabular data \citep{trees_vs_dnns1}. These importance functions are defined as follows; for additional details, see \cite{louppe} or \cite{biau_tour}. 

Given a collection of binary decision trees, $\mathcal{T}$, define
\begin{align}\label{eq:importance}
	\Phi(j) &= \frac{1}{\lvert\mathcal{T}\rvert}\sum_{T\in\mathcal{T}}\sum_{v\in T} \Delta\varphi(v)\1(j=j_v)
\end{align}
where the outer sum is over all trees $T\in\mathcal{T}$, the inner sum is over all nodes $v\in T$, $j_v$ is the feature used to split node $v$, and $\varphi(v)$ measures the impurity of $v$. The change in impurity
\begin{align*}
	\Delta\varphi(v) &= \varphi(v) - \bigg(\frac{\lvert v_L\rvert}{\lvert v\rvert}\varphi(v_L) + \frac{\lvert v_R\rvert}{\lvert v\rvert}\varphi(v_R)\bigg)
\end{align*}
is the impurity difference between $v$ and its children, $v_L$ and $v_R$. Large positive values of $\Delta\varphi(v)$ indicate that the feature $j_v$ used to split node $v$ successfully partitions the data in a manner that is consistent with the objective of the tree. 

For regression, we use the \textit{squared error loss} impurity function, $\varphi(v) = \sum_{i\in v} (Y_i - \overline Y_v)^2/\lvert v\rvert$, where $v\subseteq\{1,\dots,n\}$ is identified with the subset of samples in node $v$, and $\overline Y_v = \sum_{i\in v}Y_i/\lvert v\rvert$ is the empirical mean of the responses in $v$. For binary classification, we use the \textit{Gini index}, $\varphi(v) = 2p_0(v)p_1(v)$ where, for $a\in\{0,1\}$, $p_a(v) = \sum_{i\in v} \1(Y_i = a)/\lvert v\rvert$ is the proportion of responses in $v$ that equal $a$. These were selected because they are canonical choices and because we found little difference in results between these and other standard impurity functions. 

The importance function defined in \cref{eq:importance} can be computed for any tree ensemble method. We focus on gradient boosting \citep{gboost} and random forests \citep{rf}. As noted above, these are abbreviated \texttt{IPSSGB} and \texttt{IPSSRF} when combined with IPSS. Additional implementation details about \texttt{IPSSGB} and \texttt{IPSSRF} are in \cref{sup_sec:ipssgb,sup_sec:ipssrf}, respectively.


\subsubsection{Other importance functions.}\label{sec:other_importance_functions}


The importance function in \cref{eq:importance} is called \textit{mean decrease impurity} (MDI) because it measures the average decrease in impurity attributed to each feature over all trees. Another common importance function is \textit{mean decrease accuracy} (MDA), also known as \textit{permutation importance} \citep{louppe}. 

MDA is more general than MDI in that it can be applied to any predictive model, not just tree ensembles. Several variants of MDA exist, but the basic procedure is: (i) train the model, (ii) compute the prediction error $e$ on test data, and (iii) for each feature $j$, randomly permute the values of feature $j$ in the test data (keeping all other features unchanged), and compute the prediction error $e_j$ of the trained model on the permuted test data. The underlying idea, captured by the importance score $\Phi(j) = e_j - e$, is that permuting unimportant features should have little effect on the prediction error, while permuting important features should degrade predictive performance.

We experimented with IPSS applied to both MDI and MDA from gradient boosting and random forests. In both cases, the FDR was controlled at target levels, but IPSS with MDI consistently identified more true positives. One likely reason for this is that MDA tends to spread importances more uniformly across features than MDI, making it more difficult to distinguish between important and unimportant features \citep{esl}. Furthermore, MDI, whose importance scores are obtained during the training process itself, is more computationally efficient than MDA, which requires the additional step of evaluating the trained model on the permuted data for every feature.

The recent success of deep learning, especially on text and image data, has generated much interest in feature importance scores---for example, SHAP values \citep{shap}---that are derived from these models. Applying IPSS to such scores is a potentially interesting line of future work. However, numerous studies have shown that tree ensemble methods like \texttt{XGBoost} \citep{xgboost} are faster, easier to train, and consistently outperform deep learning methods on tabular data, especially when there are fewer than 10,000 samples \citep{trees_vs_dnns1, trees_vs_dnns2, trees_vs_dnns3}. Since our focus in this work is on tabular medical data, where the number of patients rarely exceeds several hundred, we primarily consider importance scores from tree ensemble methods given their many advantages over deep learning in this setting.


\subsubsection{Computation.}\label{sec:computation}


The IPSS subsampling procedure requires $2B$ feature importance evaluations, one evaluation on each half of the data in all $B$ iterations. This is fast for MDI scores; when combined with preselection (\cref{sup_sec:preselection}), \texttt{IPSSGB} and \texttt{IPSSRF} run in under 20 seconds on a standard laptop when there are $n=500$ samples and $p=5000$ features (\cref{tab:runtimes_reg,tab:runtimes_class}). Since this is already sufficiently fast for our purposes, we do not implement more efficient alternatives in this work.

We note, however, that IPSS is embarrassingly parallel: All iterations of the subsampling procedure can be evaluated separately and hence run simultaneously, potentially accelerating IPSS by a factor of $2B$ (typically 100 or 200). This is especially beneficial when importance scores are expensive to compute, as is often the case when they come from deep learning methods. By contrast, many feature selection methods are iterative and thus not parallelizable. For example, recursive feature elimination \citep{robust_rfe} typically removes features one at a time based on predictive performance, requiring the model to be retrained after each removal. When $p$ is large, this can require thousands of model fits, far more than the $2B$ evaluations typically required by IPSS.


\section{Simulation studies}\label{sec:simulation_studies}


In this section, we conduct two simulation studies to evaluate the performance of 14 feature selection methods when the true set of important features is known. Features in the first study are drawn from a multivariate Gaussian, and the response is generated from a nonlinear additive model. To make the simulations more realistic, in the second study we use features from real RNA sequencing (RNA-seq) data rather than generating them from known distributions, and the response is a highly randomized, nonlinear function of the important features. 


\subsection{Other methods.}\label{sec:other_methods}


We compare \texttt{IPSSGB} and \texttt{IPSSRF} to 12 feature selection methods: IPSS with $\ell^1$-regularization (\texttt{IPSSL1}, \cite{ipss}); boosting with stability selection (\texttt{SSBoost}, \cite{ssboost}); five versions of model-X knockoffs \citep{knockoffs}, namely knockoffs with generalized linear models (\texttt{KOGLM}), knockoffs with lasso (\texttt{KOL1}), knockoffs with random forests (\texttt{KORF}), knockoffs with boosted trees (\texttt{KOBT}, \cite{kobt}), and knockoffs with deep neural networks (\texttt{DeepPINK}, \cite{deeppink}); random forest hypothesis testing (\texttt{RFHT}, \cite{pitt}); \texttt{Boruta} \citep{boruta}; recursive feature elimination with gradient boosting (\texttt{RFEGB}); \texttt{Vita} \citep{vita}; and \texttt{VSURF} \citep{vsurf}. We provide a brief overview of these methods below. Additional details, including parameter settings and the software packages used to implement each method, are in \cref{sup_sec:methods}.

Four methods---\texttt{Boruta}, \texttt{RFEGB}, \texttt{Vita}, and \texttt{VSURF}---do not have theoretical false discovery control. We chose these because \cite{speiser} found that \texttt{Boruta} and \texttt{VSURF} were among the best out of $13$ random forest-based feature selection methods, and \cite{degenhardt} found that \texttt{Boruta} and \texttt{Vita} outperformed $5$ other methods in extensive comparison studies. We tested \texttt{RFEGB} so as to include a gradient boosting-based feature selection method without false discovery control.

\texttt{IPSSL1} and \texttt{SSBoost} provide theoretical E(FP) control. \texttt{IPSSL1}, a parametric version of IPSS, assumes a (generalized) linear relationship between the features and the response (\cref{eq:l1}). \texttt{SSBoost} combines gradient boosting and stability selection. We implement it assuming the $r$-concavity conditions of \cite{shah}, which are required to obtain the tightest upper bound on E(FP) of any version of stability selection other than IPSS (note that IPSS does not require $r$-concavity assumptions) \citep{ipss}.

Model-X knockoffs is a general framework for feature selection with theoretical FDR control that has attracted much attention recently, in part due to its flexibility. Like IPSS, model-X knockoffs works in high-dimensions ($p>n$), does not use $p$-values (which are challenging to compute in general), is compatible with arbitrary feature importance scores, and makes no assumptions about the relationship between the response and the features. Unlike IPSS, model-X knockoffs assumes that the joint distribution of the features is known \citep{knockoffs}. 


\subsection{Gaussian simulation design.}\label{sec:gaussian_simulations}


We perform two regression experiments and two classification experiments. The two experiments in both settings correspond to $n=250$ and $500$. All experiments consist of $100$ trials. In each trial, $n$ independent samples of $X\sim\mathcal{N}(0,\Sigma)$ are drawn from a $p=500$-dimensional, mean-zero multivariate Gaussian with a Toeplitz covariance matrix, $\Sigma$. The correlation parameter of the Toeplitz matrix is set to $\rho=0.5$; that is, $\Sigma_{jk} = 0.5^{\lvert j-k\rvert}$.

The number of important features $\lvert S\rvert$ is drawn uniformly at random from $\{5,\dots,15\}$ prior to each trial, and a new set of important features $S$ of size $\lvert S\rvert$ is randomly selected. In each experiment, the signal is $f(X) = \sum_{j\in S} \exp(-X_j^2)$. For regression, the response is $Y = f(X) + \epsilon$ where $\epsilon\sim\mathcal{N}(0,\sigma^2)$ and $\sigma^2$ is selected according to a specified signal-to-noise ratio (SNR) that is drawn uniformly at random from the interval $[0.5,2]$. For classification, we draw $Y\sim\mathrm{Bernoulli}(\pi)$ where $\pi = 1/(1 + \exp(-u f(X)))$ and the signal strength $u$ is drawn uniformly at random from the interval $[1,3]$. These sources of randomness are introduced so that the study covers a wide range of settings.

\ifthenelse{\boolean{showfigures}}{
\begin{figure*}[ht]
\includegraphics[height=.375\textheight, width=\textwidth]{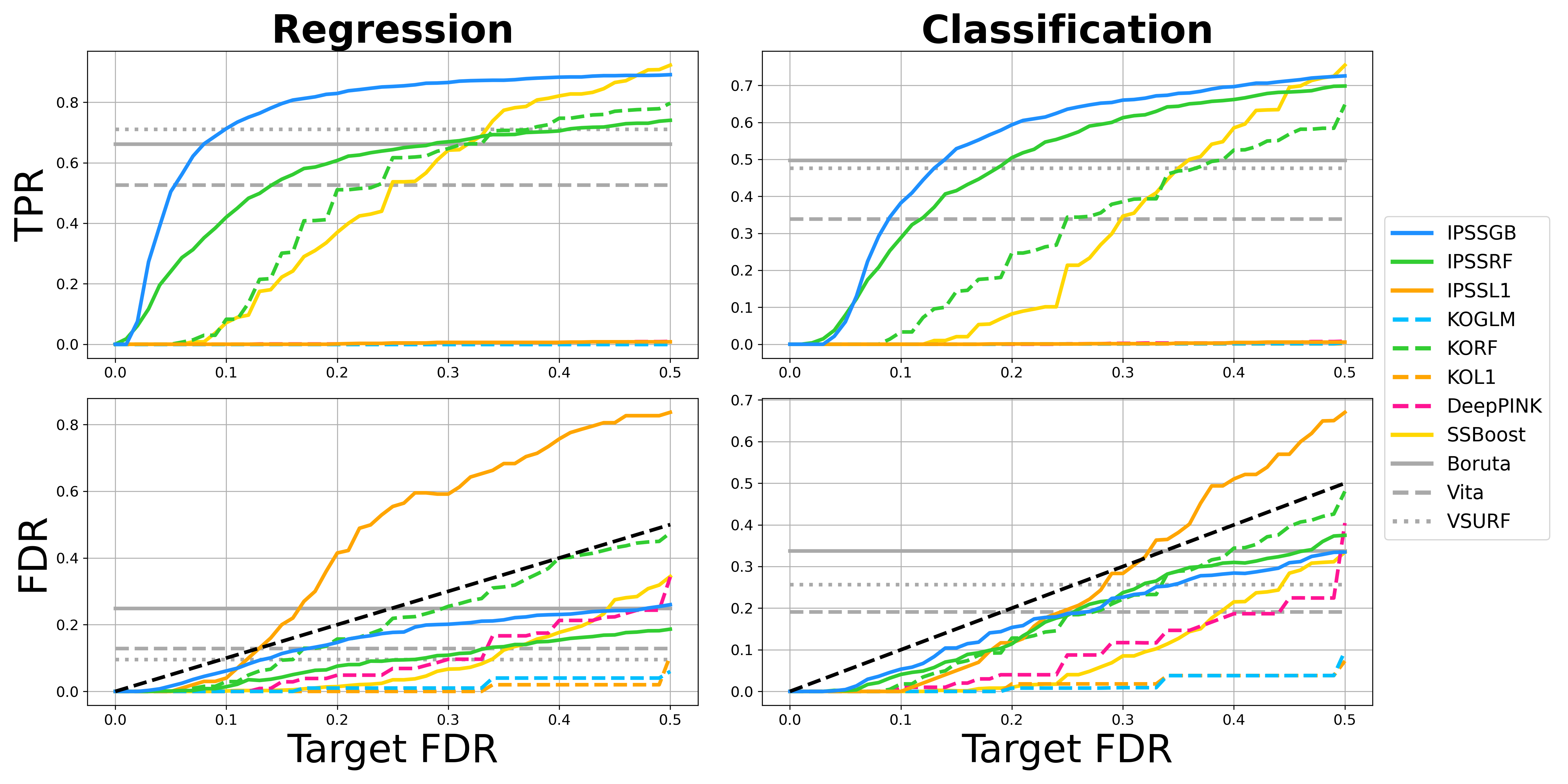}
\caption{\textit{Gaussian simulation results ($n=500$)}. First and second columns show the regression and classification results, respectively. The horizontal axis in each plot shows the target FDR. The three methods without false discovery control (gray horizontal lines) do not vary with the target FDR. The black dashed line in the FDR plots represents perfect FDR calibration, FDR = target FDR. Non-gray solid lines correspond to IPSS or stability selection-based methods. Non-gray dashed lines show methods based on model-X knockoffs.}
\label{fig:toeplitz_0.5_n500}
\end{figure*}
}{}


\subsection{RNA-seq simulation design.}\label{sec:rnaseq_simulations}


We perform three regression experiments and three classification experiments. The three experiments in both settings correspond to $p=500$, $2000$, and $5000$. All experiments consist of $100$ trials. In each trial, $n=500$ patients and $p$ genes are randomly selected from RNA-seq measurements of 6426 genes from $569$ ovarian cancer patients \citep{linkedomics}. This publicly available dataset, part of The Cancer Genome Atlas \citep{tcga}, was chosen because it is high dimensional and the features follow a variety of empirical distributions (\cref{fig:rnaseq_features}). Furthermore, the genes exhibit complex correlation structures, with maximum and average absolute pairwise correlations of approximately $0.95$ and $0.17$ after standardization, respectively.

\cref{alg:simulation} describes the simulation procedure for each individual trial, which is also illustrated in \cref{fig:simulation}. In all steps, ``randomly select" means select a parameter uniformly at random from its domain. The general outline is as follows: First, randomly select an $n$-by-$p$ submatrix $X$ of the full RNA-seq dataset and standardize its columns to have mean 0 and variance 1. Next, the number of important features $\lvert S\rvert$ is drawn uniformly at random from $\{10,\dots,30\}$ and a randomly selected subset of $\lvert S\rvert$ important features $S$ is partitioned into $G$ groups, $S_1,\dots,S_G$. A different realization of a randomized nonlinear function $f_{\theta_g}$, defined in \cref{eq:f_theta}, is applied to the standardized sum $\xi_g = (\hat\xi_g - \bar\xi_g)/\bar\sigma_g$ of the features in each group $S_g$, where $\bar\xi_g$ and $\bar\sigma_g$ are the empirical mean and standard deviation of $\hat\xi_g = \sum_{j\in S_g} X_j$. The resulting values are summed over all groups to generate a signal $\eta = \sum_{g=1}^G f_{\theta_g}(\xi_g)$, and noise is added to this signal to generate a response $Y$. This scheme produces data with highly complex interactions between features and the response, going well beyond the additive setting $Y = \sum_{j\in S} f_j(X_j) + \epsilon$.

For regression, the response is $Y = \eta + \epsilon$, where $\epsilon \sim \mathcal{N}(0,\sigma^2)$ and the variance $\sigma^2$ is selected according to a specified signal-to-noise ratio (SNR) that is drawn uniformly at random from the interval $[0.5,2]$. For classification, we draw $Y \sim \mathrm{Bernoulli}(\pi)$ where $\pi = 1/(1 + \exp(-u\eta))$ and the signal strength $u$ is drawn uniformly at random from $[1,3]$. As in the Gaussian simulation experiments, these many sources of randomness are introduced to cover a wide range of settings.


\subsection{Simulation results.}\label{sec:simulation_results}


\cref{fig:toeplitz_0.5_n500,fig:toeplitz_0.5_n250} show the results of the Gaussian simulations when $n=500$ and $250$, respectively, and \cref{fig:oc_reg} and \ref{fig:oc_class} show the results of the RNA-seq simulations for regression and classification. Runtimes for each method in each experiment are provided in \cref{tab:runtimes_reg,tab:runtimes_class}. The FDR in each plot is the average of $\mathrm{FP} / (\mathrm{TP} + \mathrm{FP})$ over all 100 trials, and the true positive rate (TPR) is the average of $\mathrm{TP} / (\mathrm{TP} + \mathrm{FN})$, where FP, TP, and FN are the number of false positives, true positives, and false negatives, respectively.

Both FDR and TPR are shown as functions of the target FDR. The black dashed line in each FDR plot represents perfect FDR calibration, $\mathrm{FDR} = \mathrm{target\ FDR}$. A method's FDR is well-controlled if its FDR lies on or below this line. The FDR and TPR for methods without false discovery control are shown as horizontal lines because they do not admit false discovery control parameters and therefore do not vary with the target FDR.

\texttt{IPSSGB} has the best performance overall. Its FDR is always well-controlled and it consistently has a much higher TPR than the other methods with false discovery control. Notably, \texttt{IPSSGB} identifies significantly more true positives than other methods with false discovery control at lower target FDRs. For example, in the regression results in \cref{fig:toeplitz_0.5_n500}, \texttt{IPSSGB} identifies 70\% of important features when the target FDR is 0.1, while \texttt{IPSSRF} identifies 40\% and the remaining methods identify close to none. \texttt{IPSSGB}'s TPR even surpasses the TPR of methods without error control in almost all experiments despite having far fewer false positives. With an average runtime of less than 15 seconds in all experiments, \texttt{IPSSGB} is also one of the fastest methods.

Among the other methods with theoretical error control, \texttt{IPSSRF} performs well in terms of identifying true positives while controlling false positives, though not as well as \texttt{IPSSGB}. \texttt{IPSSL1}, whose parametric assumptions are violated, performs poorly, failing to control false positives at target FDR levels. It also identifies essentially no true positives in the Gaussian experiments. \texttt{SSBoost} is overly conservative, undershooting the target FDR at the expense of identifying few true positives. This is partly due to the weakness of the efp scores used by \texttt{SSBoost} relative to those used by IPSS \citep{ipss}.

All versions of model-X knockoffs underperform \texttt{IPSSGB} and \texttt{IPSSRF} in all experiments. This is even true in the Gaussian setting, where the known joint distribution of the features was used to implement these methods (recall that model-X knockoffs requires knowledge of the joint distribution, while IPSS does not). With the exception of \texttt{KOL} and \texttt{KORF} in the $p=5000$ and, to a lesser extent, $p=2000$ RNA-seq experiments, all of these methods control the FDR at target levels. \texttt{KORF} is the only knockoffs-based method that identifies essentially any true positives in the Gaussian experiments. \texttt{DeepPINK}, which combines knockoffs with deep neural networks, has the lowest TPR out of every method in almost all experiments. This agrees with our earlier observation that deep learning typically underperforms tree-based methods on tabular data and requires tens of thousands of samples for competitive performance \citep{trees_vs_dnns2}.

The absence of error control for \texttt{Boruta} and \texttt{Vita} is clearly apparent in \cref{fig:oc_reg,fig:oc_class}. Both methods usually have FDRs over 0.75, far surpassing other methods. Despite this, \texttt{Boruta} and \texttt{Vita} almost always identify fewer true positives than \texttt{IPSSGB} when the target FDR is greater than 0.2 or, in some cases, even 0.1. \texttt{Boruta} and \texttt{Vita} are also slower than other methods, and this disparity grows with the number of features (\cref{tab:runtimes_reg,tab:runtimes_class}).

\texttt{VSURF} usually has a lower FDR than \texttt{Boruta} and \texttt{Vita}, but still underperforms \texttt{IPSSGB}. Its excessive runtimes prevented us from including it in the $p=2000$ and $5000$ RNA-seq experiments. Several other methods are also omitted, namely \texttt{KOBT}, \texttt{RFEGB}, and \texttt{RFHT}. Briefly, \texttt{KOBT} far exceeded target FDRs despite extensive tuning, while \texttt{RFEGB} and \texttt{RFHT} had FDRs over $0.8$ and extremely long runtimes. For details, see \cref{sup_sec:simulations,sup_sec:methods}.

\ifthenelse{\boolean{showfigures}}{
\begin{figure*}[ht]
\includegraphics[height=.325\textheight, width=\textwidth]{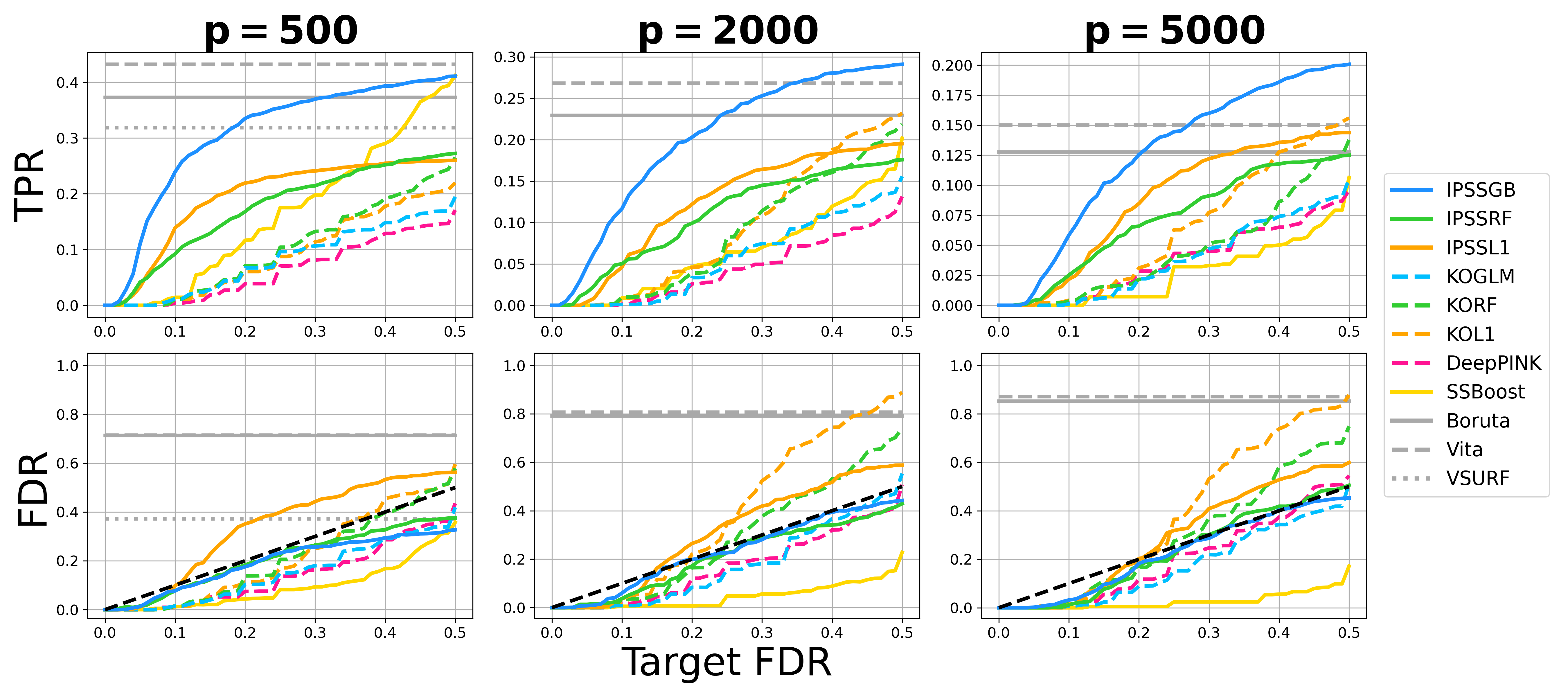}
\caption{\textit{RNA-seq simulation results (regression)}. First, second, and third columns correspond to the $p=500$, $2000$, and $5000$ experiments, respectively. The horizontal axis in each plot shows the target FDR. Methods without false discovery control (gray horizontal lines) do not vary with the target FDR. The black dashed line in the FDR plots represents perfect FDR calibration, FDR = target FDR. Non-gray solid lines correspond to IPSS or stability selection-based methods. Non-gray dashed lines show methods based on model-X knockoffs.}
\label{fig:oc_reg}
\end{figure*}
}{}


\section{Cancer studies}\label{sec:cancer}


We study ovarian cancer and glioma (a type of brain cancer). Both studies include multiple substudies in which the features are either genes, measured by RNA-seq, or microRNAs (miRNAs). The response variables are either \textit{prognosis} (whether the patient was alive at last follow-up), \textit{tumor purity} (the proportion of cancerous cells in a tissue sample), or the expression level of a particular gene or miRNA. All data are from The Cancer Genome Atlas \citep{tcga} and were downloaded from LinkedOmics \citep{linkedomics}. Additional details and results are in \cref{sup_sec:cancer}.

We assess feature selection performance by (i) performing literature searches and (ii) assessing predictive performance using cross-validation (see \cref{sup_sec:lit_search,sup_sec:cv_study}, respectively). The literature search results show that \texttt{IPSSGB} and \texttt{IPSSRF} consistently identify more known important features at lower target FDRs than other methods. In \cref{tab:ovarian_mirna_status}, for example, \texttt{IPSSGB} and \texttt{IPSSRF} identify 8 and 6 miRNAs, respectively, all but one of which have been implicated in ovarian cancer prognosis. In contrast, \texttt{IPSSL1} identifies 4 miRNAs, missing the three most significant ones, while \texttt{KOGLM}, \texttt{KORF}, \texttt{KOL1}, \texttt{DeepPINK}, and \texttt{SSBoost} select no miRNAs at all, even when the target FDR is $0.5$. 

In the RNA-seq and glioma prognosis study (\cref{tab:glioma_rnaseq_status}), only \texttt{IPSSGB} identifies the key oncogene FOXM1 \citep{foxm1}, and only \texttt{IPSSGB}, \texttt{IPSSRF}, and \texttt{KORF} identify WEE1, which is also known to play a significant role in glioma outcomes \citep{wee1}. Furthermore, \texttt{IPSSGB} and \texttt{IPSSRF} are more confident in their selections, assigning WEE1 $q$-values of 0.10 and 0.06, respectively, while \texttt{KORF} does not select WEE1 until the target FDR is reduced to 0.44. More generally, \cref{tab:glioma_rnaseq_status} shows that \texttt{IPSSGB}, \texttt{IPSSRF}, and, to a lesser extent, \texttt{IPSSL1}, tend to identify genes supported by the glioma literature while avoiding genes with little or no known connection. In contrast, \texttt{KORF} and especially \texttt{KOL1} select many genes with limited or no literature support, while \texttt{KOGLM}, \texttt{DeepPINK}, and \texttt{SSBoost} select no genes at all.

Briefly, our cross-validation (CV) studies proceed as follows. In each of 20 CV steps, one group of patients is set aside (the test set), and a set of features is selected by each method using the data in the remaining groups (the training set). Next, for each method we construct three predictive models---a linear model, a random forest model, and a gradient boosting model---using only the features selected by that method on the training data. Each model is then used to predict responses from the test set, and the smallest of the three prediction errors is recorded (we use mean squared error for regression and $1 - \mathrm{accuracy}$ for classification). All three models are implemented so that no method has an inherent advantage over another. For example, features selected by \texttt{IPSSL1} may be better suited to linear model predictions than those selected by \texttt{IPSSGB}, while those selected by \texttt{IPSSGB} may be better suited to gradient boosting predictions than the ones selected by \texttt{IPSSL1}.

\cref{fig:cv_study_glioma_rnaseq_status_main} shows the results from our RNA-seq and glioma prognosis CV study. On average, the top 20 genes selected by \texttt{IPSSGB} yield the same prediction error as the full model that uses all $10,058$ genes. \texttt{IPSSRF} and \texttt{IPSSL1} achieve the smallest prediction errors overall, and only use 10 to 20 genes to do so. \texttt{KORF} and \texttt{KOL1} have higher predictive errors despite using more than 40 selected genes, and \texttt{Boruta}, which selects over 100 genes, has an average error similar to that of the full model. \texttt{Vita} (not shown) selects over 500 genes on average and has an average prediction error of 0.22, and \texttt{DeepPINK} and \texttt{SSBoost} select no genes at all, even when the target FDR is 0.5.

\ifthenelse{\boolean{showfigures}}{
\begin{figure}[ht]
\includegraphics[width=\textwidth]{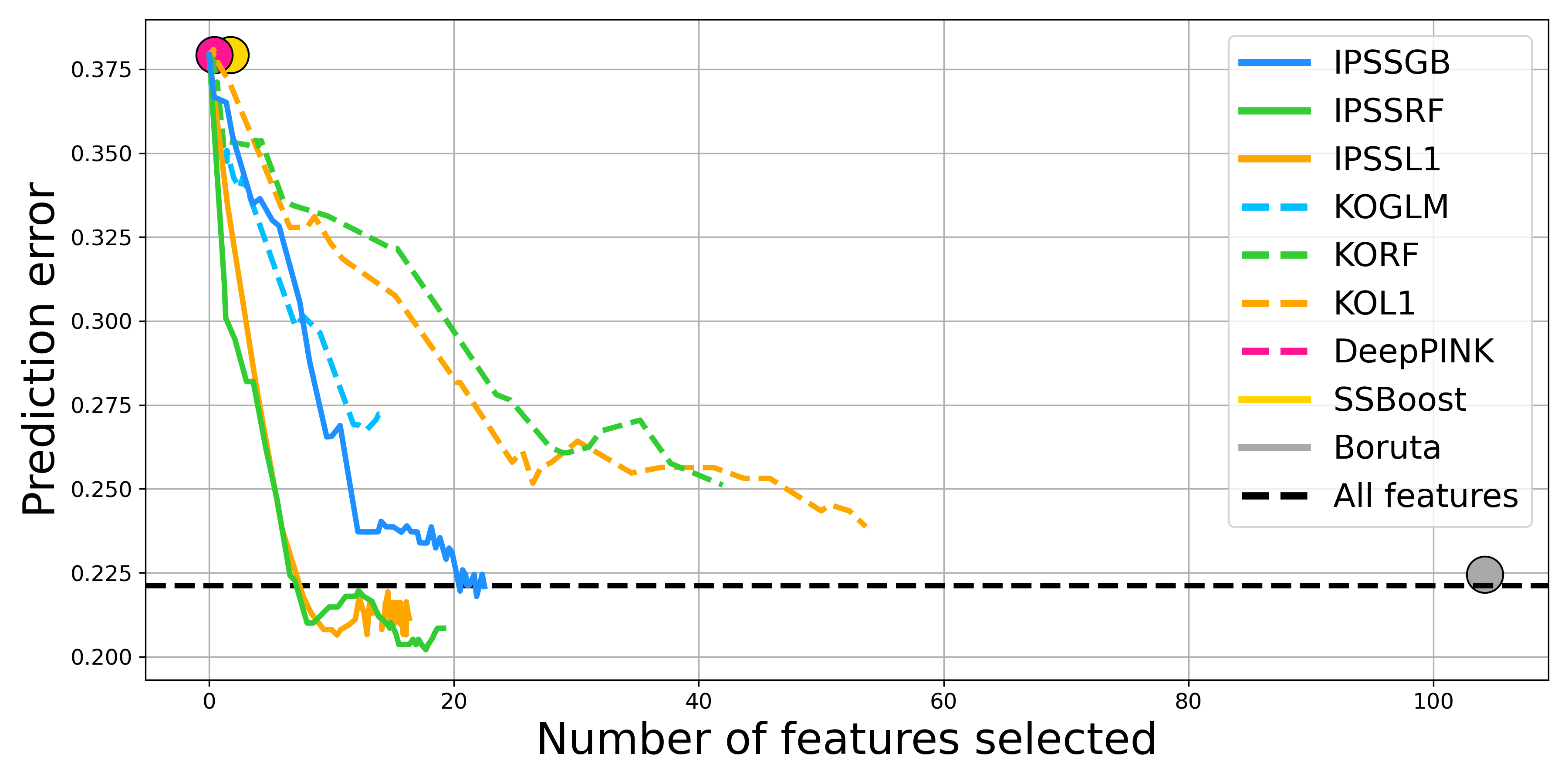}
\caption{\textit{RNA-seq and glioma prognosis}. The horizontal and vertical axes show the average number of genes selected and the average prediction error over the 20 cross-validation steps, respectively. Curves for each method are obtained by varying the target FDR between 0 and 0.5. \texttt{DeepPINK} and \texttt{SSBoost} selected no genes and are therefore shown by single points rather than curves. \texttt{Boruta} is also indicated by a single point because it does not have FDR control parameters. The dashed black line shows the average error when using all $p=10,058$ genes to predict prognosis.}
\label{fig:cv_study_glioma_rnaseq_status_main}
\end{figure}
}{}


\section{Discussion}\label{sec:discussion}


We have demonstrated that \texttt{IPSSRF} and \texttt{IPSSGB} achieve superior results in terms of false positive control, true positive detection, and computation time. More broadly, IPSS for thresholding is a general framework whose theory and implementation apply to arbitrary importance scores. For instance, examples of other scores include $p$-values (with smaller $p$-values indicating greater importance), Shapley values (used to quantify the contribution of individual features to neural networks and other machine learning models), and loadings in principal components analysis (which quantify the contribution of each feature to a given principal component). The main practical limitation to consider is the cost of computing the relevant importance scores, since IPSS must compute these scores on multiple subsamples of the data.

We have also introduced efp scores and shown that, in addition to controlling E(FP), they can be used to control the FDR and estimate $q$-values. \cite{storey_pfdr} showed that $q$-values admit a Bayesian interpretation, suggesting a link between IPSS and Bayesian feature selection that could be an interesting line of future work.

Finally, a more ambitious goal is to extend IPSS to unsupervised feature selection problems (that is, feature selection when there is no response variable) and non-iid data. The PCA-based scores mentioned above provide at least one way to apply IPSS in an unsupervised setting. Developing a rigorous approach to IPSS for non-iid data could provide novel methods for nonparametric feature selection with false discovery control for networks and spatially or temporally-indexed data.


\section*{Data and code availability}


All code and data used in this work are available on GitHub 
(\url{https://github.com/omelikechi/ipss_bioinformatics}) and permanently archived on Zenodo 
(\url{https://doi.org/10.5281/zenodo.15335289}). A Python package for implementing IPSS is available on GitHub 
(\url{https://github.com/omelikechi/ipss}) and PyPI (\url{https://pypi.org/project/ipss/}). 
An R implementation of IPSS is also available on GitHub 
(\url{https://github.com/omelikechi/ipssR}).


\section*{Acknowledgements}


D.B.D. and O.M. were supported in part by funding
from Merck \& Co. and the National Institutes of Health (NIH) grant R01ES035625. J.W.M. and O.M. were supported in part by the Collaborative Center for X-linked Dystonia Parkinsonism (CCXDP). J.W.M. was supported in part by the National Institutes of Health (NIH) grant R01CA240299.


\bibliographystyle{abbrvnat}
\bibliography{refs}

\begin{thebibliography}{50}
\providecommand{\natexlab}[1]{#1}
\providecommand{\url}[1]{\texttt{#1}}
\expandafter\ifx\csname urlstyle\endcsname\relax
  \providecommand{\doi}[1]{doi: #1}\else
  \providecommand{\doi}{doi: \begingroup \urlstyle{rm}\Url}\fi

\bibitem[Biau and Scornet(2016)]{biau_tour}
G.~Biau and E.~Scornet.
\newblock A random forest guided tour.
\newblock \emph{Test}, 25:\penalty0 197--227, 2016.

\bibitem[Breiman(2001)]{rf}
L.~Breiman.
\newblock Random forests.
\newblock \emph{Machine Learning}, 45:\penalty0 5--32, 2001.

\bibitem[Candes et~al.(2018)Candes, Fan, Janson, and Lv]{knockoffs}
E.~Candes, Y.~Fan, L.~Janson, and J.~Lv.
\newblock Panning for gold: ‘model-x’ knockoffs for high dimensional
  controlled variable selection.
\newblock \emph{Journal of the Royal Statistical Society Series B: Statistical
  Methodology}, 80\penalty0 (3):\penalty0 551--577, 2018.

\bibitem[Chen and Guestrin(2016)]{xgboost}
T.~Chen and C.~Guestrin.
\newblock Xgboost: {A} scalable tree boosting system.
\newblock In \emph{Proceedings of the 22nd ACM SIGKDD International Conference
  on Knowledge Discovery and Data Mining}, pages 785--794, 2016.

\bibitem[Coleman et~al.(2022)Coleman, Peng, and Mentch]{pitt}
T.~Coleman, W.~Peng, and L.~Mentch.
\newblock Scalable and efficient hypothesis testing with random forests.
\newblock \emph{Journal of Machine Learning Research}, 23\penalty0
  (170):\penalty0 1--35, 2022.

\bibitem[Degenhardt et~al.(2019)Degenhardt, Seifert, and Szymczak]{degenhardt}
F.~Degenhardt, S.~Seifert, and S.~Szymczak.
\newblock Evaluation of variable selection methods for random forests and omics
  data sets.
\newblock \emph{Briefings in Bioinformatics}, 20\penalty0 (2):\penalty0
  492--503, 2019.

\bibitem[Dou et~al.(2020)Dou, Chen, Lu, Qiu, and Zhang]{mir342}
Y.~Dou, F.~Chen, Y.~Lu, H.~Qiu, and H.~Zhang.
\newblock {Effects of Wnt/$\beta$-catenin signal pathway regulated by
  miR-342-5p targeting CBX2 on proliferation, metastasis and invasion of
  ovarian cancer cells}.
\newblock \emph{Cancer Management and Research}, pages 3783--3794, 2020.

\bibitem[{Europe PMC Consortium}(2015)]{europePMC}
{Europe PMC Consortium}.
\newblock {Europe PMC: a full-text literature database for the life sciences
  and platform for innovation}.
\newblock \emph{Nucleic Acids Research}, 43\penalty0 (D1):\penalty0
  D1042--D1048, 2015.

\bibitem[Fayaz et~al.(2022)Fayaz, Zaman, Kaul, and Butt]{trees_vs_dnns3}
S.~A. Fayaz, M.~Zaman, S.~Kaul, and M.~A. Butt.
\newblock Is deep learning on tabular data enough? {A}n assessment.
\newblock \emph{International Journal of Advanced Computer Science and
  Applications}, 13\penalty0 (4):\penalty0 466--473, 2022.

\bibitem[Friedman(2001)]{gboost}
J.~H. Friedman.
\newblock Greedy function approximation: a gradient boosting machine.
\newblock \emph{Annals of Statistics}, pages 1189--1232, 2001.

\bibitem[Friedman et~al.(2010)Friedman, Hastie, and
  Tibshirani]{logistic_regression}
J.~H. Friedman, T.~Hastie, and R.~Tibshirani.
\newblock Regularization paths for generalized linear models via coordinate
  descent.
\newblock \emph{Journal of Statistical Software}, 33:\penalty0 1--22, 2010.

\bibitem[Fu et~al.(2012)Fu, Tian, Zhang, Chen, and Hao]{mir93_1}
X.~Fu, J.~Tian, L.~Zhang, Y.~Chen, and Q.~Hao.
\newblock {Involvement of microRNA-93, a new regulator of PTEN/Akt signaling
  pathway, in regulation of chemotherapeutic drug cisplatin chemosensitivity in
  ovarian cancer cells}.
\newblock \emph{{FEBS Letters}}, 586\penalty0 (9):\penalty0 1279--1286, 2012.

\bibitem[Genuer et~al.(2010)Genuer, Poggi, and Tuleau-Malot]{vsurf}
R.~Genuer, J.-M. Poggi, and C.~Tuleau-Malot.
\newblock Variable selection using random forests.
\newblock \emph{Pattern Recognition Letters}, 31\penalty0 (14):\penalty0
  2225--2236, 2010.

\bibitem[Ghafouri-Fard et~al.(2022)Ghafouri-Fard, Khoshbakht, Hussen, Sarfaraz,
  Taheri, and Ayatollahi]{mir1270}
S.~Ghafouri-Fard, T.~Khoshbakht, B.~M. Hussen, S.~Sarfaraz, M.~Taheri, and
  S.~A. Ayatollahi.
\newblock {Circ\_CDR1as: A circular RNA with roles in the carcinogenesis}.
\newblock \emph{Pathology-Research and Practice}, 236:\penalty0 153968, 2022.

\bibitem[Grinsztajn et~al.(2022)Grinsztajn, Oyallon, and
  Varoquaux]{trees_vs_dnns2}
L.~Grinsztajn, E.~Oyallon, and G.~Varoquaux.
\newblock Why do tree-based models still outperform deep learning on typical
  tabular data?
\newblock \emph{Advances in Neural Information Processing Systems},
  35:\penalty0 507--520, 2022.

\bibitem[Hastie et~al.(2009)Hastie, Tibshirani, and Friedman]{esl}
T.~Hastie, R.~Tibshirani, and J.~H. Friedman.
\newblock \emph{{The Elements of Statistical Learning: Data Mining, Inference,
  and Prediction}}, volume~2.
\newblock Springer, 2009.

\bibitem[Hofner and Hothorn(2017)]{stabs}
B.~Hofner and T.~Hothorn.
\newblock {stabs: Stability selection with error control}.
\newblock \emph{R package version 0.6-3}, 2017.

\bibitem[Hofner et~al.(2014)Hofner, Mayr, Robinzonov, and Schmid]{mboost}
B.~Hofner, A.~Mayr, N.~Robinzonov, and M.~Schmid.
\newblock Model-based boosting in {R}: a hands-on tutorial using the {R}
  package mboost.
\newblock \emph{Computational Statistics}, 29:\penalty0 3--35, 2014.

\bibitem[Hofner et~al.(2015)Hofner, Boccuto, and G{\"o}ker]{ssboost}
B.~Hofner, L.~Boccuto, and M.~G{\"o}ker.
\newblock Controlling false discoveries in high-dimensional situations:
  boosting with stability selection.
\newblock \emph{BMC Bioinformatics}, 16:\penalty0 1--17, 2015.

\bibitem[Janitza et~al.(2018)Janitza, Celik, and Boulesteix]{vita}
S.~Janitza, E.~Celik, and A.-L. Boulesteix.
\newblock A computationally fast variable importance test for random forests
  for high-dimensional data.
\newblock \emph{Advances in Data Analysis and Classification}, 12\penalty0
  (4):\penalty0 885--915, 2018.

\bibitem[Jiang et~al.(2021)Jiang, Li, and Motsinger-Reif]{kobt}
T.~Jiang, Y.~Li, and A.~A. Motsinger-Reif.
\newblock Knockoff boosted tree for model-free variable selection.
\newblock \emph{Bioinformatics}, 37\penalty0 (7):\penalty0 976--983, 2021.

\bibitem[Jin et~al.(2014)Jin, Yang, Ye, Xu, and Hua]{mir150_1}
M.~Jin, Z.~Yang, W.~Ye, H.~Xu, and X.~Hua.
\newblock {MicroRNA-150 predicts a favorable prognosis in patients with
  epithelial ovarian cancer, and inhibits cell invasion and metastasis by
  suppressing transcriptional repressor ZEB1}.
\newblock \emph{PloS One}, 9\penalty0 (8):\penalty0 e103965, 2014.

\bibitem[Kandettu et~al.(2022)Kandettu, Adiga, Devi, Suresh, Chakrabarty,
  Radhakrishnan, and Kabekkodu]{mir1-2}
A.~Kandettu, D.~Adiga, V.~Devi, P.~S. Suresh, S.~Chakrabarty, R.~Radhakrishnan,
  and S.~P. Kabekkodu.
\newblock {Deregulated miRNA clusters in ovarian cancer: Imperative
  implications in personalized medicine}.
\newblock \emph{Genes \& Diseases}, 9\penalty0 (6):\penalty0 1443--1465, 2022.

\bibitem[Kim et~al.(2017)Kim, Jeong, Park, Kim, Heo, Kang, Kim, and
  An]{mir150_2}
T.~H. Kim, J.-Y. Jeong, J.-Y. Park, S.-W. Kim, J.~H. Heo, H.~Kang, G.~Kim, and
  H.~J. An.
\newblock {miR-150 enhances apoptotic and anti-tumor effects of paclitaxel in
  paclitaxel-resistant ovarian cancer cells by targeting Notch3}.
\newblock \emph{Oncotarget}, 8\penalty0 (42):\penalty0 72788, 2017.

\bibitem[Kursa et~al.(2010)Kursa, Jankowski, and Rudnicki]{boruta}
M.~B. Kursa, A.~Jankowski, and W.~R. Rudnicki.
\newblock Boruta--a system for feature selection.
\newblock \emph{Fundamenta Informaticae}, 101\penalty0 (4):\penalty0 271--285,
  2010.

\bibitem[Lee et~al.(2012)Lee, Park, Deftereos, Morihara, Stern, Hawes, Swisher,
  Kiviat, and Feng]{mir30d_2}
H.~Lee, C.~S. Park, G.~Deftereos, J.~Morihara, J.~E. Stern, S.~E. Hawes,
  E.~Swisher, N.~B. Kiviat, and Q.~Feng.
\newblock {MicroRNA expression in ovarian carcinoma and its correlation with
  clinicopathological features}.
\newblock \emph{World Journal of Surgical Oncology}, 10:\penalty0 1--10, 2012.

\bibitem[Li et~al.(2022)Li, Ching, and Liu]{robust_rfe}
L.~Li, W.-K. Ching, and Z.-P. Liu.
\newblock Robust biomarker screening from gene expression data by stable
  machine learning-recursive feature elimination methods.
\newblock \emph{Computational Biology and Chemistry}, 100:\penalty0 107747,
  2022.

\bibitem[Liu et~al.(2019)Liu, Zhang, and Yang]{mir96_1}
B.~Liu, J.~Zhang, and D.~Yang.
\newblock {miR-96-5p promotes the proliferation and migration of ovarian cancer
  cells by suppressing Caveolae1}.
\newblock \emph{Journal of Ovarian Research}, 12:\penalty0 1--9, 2019.

\bibitem[Louppe et~al.(2013)Louppe, Wehenkel, Sutera, and Geurts]{louppe}
G.~Louppe, L.~Wehenkel, A.~Sutera, and P.~Geurts.
\newblock Understanding variable importances in forests of randomized trees.
\newblock \emph{Advances in Neural Information Processing Systems}, 26, 2013.

\bibitem[Lu et~al.(2018)Lu, Fan, Lv, and Stafford~Noble]{deeppink}
Y.~Lu, Y.~Fan, J.~Lv, and W.~Stafford~Noble.
\newblock {DeepPINK}: reproducible feature selection in deep neural networks.
\newblock \emph{Advances in Neural Information Processing Systems}, 31, 2018.

\bibitem[Lundberg and Lee(2017)]{shap}
S.~M. Lundberg and S.-I. Lee.
\newblock A unified approach to interpreting model predictions.
\newblock \emph{Advances in Neural Information Processing Systems}, 30, 2017.

\bibitem[Meinshausen and B{\"u}hlmann(2010)]{mb}
N.~Meinshausen and P.~B{\"u}hlmann.
\newblock Stability selection.
\newblock \emph{Journal of the Royal Statistical Society Series B: Statistical
  Methodology}, 72\penalty0 (4):\penalty0 417--473, 2010.

\bibitem[Melikechi and Miller(2025)]{ipss}
O.~Melikechi and J.~W. Miller.
\newblock Integrated path stability selection.
\newblock \emph{Journal of the American Statistical Association}, 2025.
\newblock Published online.
  \url{https://doi.org/10.1080/01621459.2025.2525589}.

\bibitem[Meng et~al.(2015)Meng, Joosse, M{\"u}ller, Trillsch, Milde-Langosch,
  Mahner, Geffken, Pantel, and Schwarzenbach]{mir93_2}
X.~Meng, S.~A. Joosse, V.~M{\"u}ller, F.~Trillsch, K.~Milde-Langosch,
  S.~Mahner, M.~Geffken, K.~Pantel, and H.~Schwarzenbach.
\newblock {Diagnostic and prognostic potential of serum miR-7, miR-16, miR-25,
  miR-93, miR-182, miR-376a and miR-429 in ovarian cancer patients}.
\newblock \emph{British Journal of Cancer}, 113\penalty0 (9):\penalty0
  1358--1366, 2015.

\bibitem[Music et~al.(2016)Music, Dahlrot, Hermansen, Hjelmborg, de~Stricker,
  Hansen, and Kristensen]{wee1}
D.~Music, R.~H. Dahlrot, S.~K. Hermansen, J.~Hjelmborg, K.~de~Stricker,
  S.~Hansen, and B.~W. Kristensen.
\newblock {Expression and prognostic value of the WEE1 kinase in gliomas}.
\newblock \emph{Journal of neuro-oncology}, 127:\penalty0 381--389, 2016.

\bibitem[Nogueira et~al.(2018)Nogueira, Sechidis, and Brown]{nogueira}
S.~Nogueira, K.~Sechidis, and G.~Brown.
\newblock On the stability of feature selection algorithms.
\newblock \emph{Journal of Machine Learning Research}, 18\penalty0
  (174):\penalty0 1--54, 2018.

\bibitem[Pedregosa et~al.(2011)Pedregosa, Varoquaux, Gramfort, Michel, Thirion,
  Grisel, Blondel, Prettenhofer, Weiss, Dubourg, et~al.]{sklearn}
F.~Pedregosa, G.~Varoquaux, A.~Gramfort, V.~Michel, B.~Thirion, O.~Grisel,
  M.~Blondel, P.~Prettenhofer, R.~Weiss, V.~Dubourg, et~al.
\newblock Scikit-learn: {M}achine learning in python.
\newblock \emph{Journal of Machine Learning Research}, 12:\penalty0 2825--2830,
  2011.

\bibitem[Raychaudhuri and Park(2011)]{foxm1}
P.~Raychaudhuri and H.~J. Park.
\newblock {FoxM1: a master regulator of tumor metastasis}.
\newblock \emph{Cancer research}, 71\penalty0 (13):\penalty0 4329--4333, 2011.

\bibitem[Shah and Samworth(2013)]{shah}
R.~D. Shah and R.~J. Samworth.
\newblock Variable selection with error control: another look at stability
  selection.
\newblock \emph{Journal of the Royal Statistical Society Series B: Statistical
  Methodology}, 75\penalty0 (1):\penalty0 55--80, 2013.

\bibitem[Shwartz-Ziv and Armon(2022)]{trees_vs_dnns1}
R.~Shwartz-Ziv and A.~Armon.
\newblock Tabular data: Deep learning is not all you need.
\newblock \emph{Information Fusion}, 81:\penalty0 84--90, 2022.

\bibitem[Speiser et~al.(2019)Speiser, Miller, Tooze, and Ip]{speiser}
J.~L. Speiser, M.~E. Miller, J.~Tooze, and E.~Ip.
\newblock A comparison of random forest variable selection methods for
  classification prediction modeling.
\newblock \emph{Expert Systems with Applications}, 134:\penalty0 93--101, 2019.

\bibitem[Storey(2003)]{storey_pfdr}
J.~D. Storey.
\newblock The positive false discovery rate: a bayesian interpretation and the
  q-value.
\newblock \emph{Annals of Statistics}, 31\penalty0 (6):\penalty0 2013--2035,
  2003.

\bibitem[Storey and Tibshirani(2003)]{storey_gwas}
J.~D. Storey and R.~Tibshirani.
\newblock Statistical significance for genomewide studies.
\newblock \emph{Proceedings of the National Academy of Sciences}, 100\penalty0
  (16):\penalty0 9440--9445, 2003.

\bibitem[Theng and Bhoyar(2024)]{survey}
D.~Theng and K.~K. Bhoyar.
\newblock Feature selection techniques for machine learning: a survey of more
  than two decades of research.
\newblock \emph{Knowledge and Information Systems}, 66\penalty0 (3):\penalty0
  1575--1637, 2024.

\bibitem[Tibshirani(1996)]{lasso}
R.~Tibshirani.
\newblock Regression shrinkage and selection via the lasso.
\newblock \emph{Journal of the Royal Statistical Society Series B: Statistical
  Methodology}, 58\penalty0 (1):\penalty0 267--288, 1996.

\bibitem[Vasaikar et~al.(2018)Vasaikar, Straub, Wang, and Zhang]{linkedomics}
S.~V. Vasaikar, P.~Straub, J.~Wang, and B.~Zhang.
\newblock Linkedomics: analyzing multi-omics data within and across 32 cancer
  types.
\newblock \emph{Nucleic Acids Research}, 46\penalty0 (D1):\penalty0 D956--D963,
  2018.

\bibitem[Weinstein et~al.(2013)Weinstein, Collisson, Mills, Shaw, Ozenberger,
  Ellrott, Shmulevich, Sander, and Stuart]{tcga}
J.~N. Weinstein, E.~A. Collisson, G.~B. Mills, K.~R. Shaw, B.~A. Ozenberger,
  K.~Ellrott, I.~Shmulevich, C.~Sander, and J.~M. Stuart.
\newblock The cancer genome atlas pan-cancer analysis project.
\newblock \emph{Nature Genetics}, 45\penalty0 (10):\penalty0 1113--1120, 2013.

\bibitem[Yang et~al.(2020)Yang, Zhang, and Bi]{mir96_2}
N.~Yang, Q.~Zhang, and X.-J. Bi.
\newblock {miRNA-96 accelerates the malignant progression of ovarian cancer via
  targeting FOXO3a.}
\newblock \emph{European Review for Medical \& Pharmacological Sciences},
  24\penalty0 (1), 2020.

\bibitem[Ye et~al.(2015)Ye, Zhao, Li, Chen, and Li]{mir30d_1}
Z.~Ye, L.~Zhao, J.~Li, W.~Chen, and X.~Li.
\newblock {miR-30d blocked transforming growth factor $\beta$1--induced
  epithelial-mesenchymal transition by targeting snail in ovarian cancer
  cells}.
\newblock \emph{International Journal of Gynecologic Cancer}, 25\penalty0 (9),
  2015.

\bibitem[Yu and Gao(2020)]{mir1301}
J.-L. Yu and X.~Gao.
\newblock {MicroRNA 1301 inhibits cisplatin resistance in human ovarian cancer
  cells by regulating EMT and autophagy.}
\newblock \emph{European Review for Medical \& Pharmacological Sciences},
  24\penalty0 (4), 2020.

\end{thebibliography}


\clearpage

\setcounter{page}{1}
\setcounter{section}{0}
\setcounter{table}{0}
\setcounter{figure}{0}
\renewcommand{\theHsection}{SIsection.\arabic{section}}
\renewcommand{\theHtable}{SItable.\arabic{table}}
\renewcommand{\theHfigure}{SIfigure.\arabic{figure}}
\renewcommand{\thepage}{S\arabic{page}}  
\renewcommand{\thesection}{S\arabic{section}}   
\renewcommand{\thetable}{S\arabic{table}}   
\renewcommand{\thefigure}{S\arabic{figure}}
\renewcommand{\thealgorithm}{S\arabic{algorithm}}

\begin{center}
{\Large\textbf{{\LARGE S}UPPLEMENTARY MATERIAL}}
\end{center}

We provide further details about IPSS (\cref{sup_sec:ipss}), describe the feature selection methods considered in this work (\cref{sup_sec:methods}), and present results from our simulation studies (\cref{sup_sec:simulations}), cancer studies (\cref{sup_sec:cancer}), and sensitivity analyses of the IPSS parameters (\cref{sup_sec:sensitivity}).


\vspace*{1em}
\section{IPSS details}\label{sup_sec:ipss}


We provide an algorithm that implements integrated path stability selection (IPSS) for feature importance scores (\cref{sup_sec:ipss_algorithm}), elaborate on the theory of IPSS and its connection to efp scores (\cref{sup_sec:theory}), and describe the IPSS parameters in greater detail (\cref{sup_sec:parameters}).


\subsection{Algorithm.}\label{sup_sec:ipss_algorithm}


\cref{alg:ipss}, discussed in \cref{sec:ipss}, implements IPSS for feature importance scores. The number of grid points used to evaluate the integrals in \cref{alg:ipss} is always $K=100$. Like many of the other IPSS parameters, $K$ is inconsequential provided it is sufficiently large; in our experience, values greater than 25 suffice. This is because the function $f(x)=(2x-1)^3\1(x\geq 0.5)$, the paths $\lambda\mapsto\hat\pi_j(\lambda)$ (which are monotonically increasing functions of $\lambda$), the quantity $\mathcal{I}(\Lambda)$, and the family of measures $\mu_\delta(d\lambda) = z^{-1}_\delta\lambda^{-\delta}d\lambda$ are all very numerically stable.

\begin{algorithm}[!ht]
\caption{Integrated path stability selection for feature importance scores}\label{alg:ipss}
\begin{algorithmic}[1]
\Require{Data $Z_{1:n}$, importance function $\Phi$, number of grid points $K$ and iterations $B$, probability measure $\mu$, function $f$ (default $f(x)=(2x-1)^3\1(x\geq 0.5)$), and cutoff $C$ (default $C=0.05$).}
\State (Optional) Preselect features, as described in \cref{sup_sec:preselection}.
\For{$b=1,\ldots,B$}
\State Randomly select $A_{2b-1},A_{2b}\subseteq\{1,\ldots,n\}$ with $A_{2b-1}\cap A_{2b}=\varnothing$ and $\lvert A_{2b-1}\rvert=\lvert A_{2b}\rvert=\lfloor n/2\rfloor$.
\State Evaluate $\Phi_{Z_{A_{2b-1}}}(j)$ and $\Phi_{Z_{A_{2b}}}(j)$ for $j=1,\dots,p$.
\EndFor
\State\label{step:lmax} Set $\lmax = \max\{\Phi_{Z_{A_b}}(j) : 1\leq b\leq 2B,\, 1\leq j\leq p\}$.
\State Define a $\lambda$ grid with upper bound $\lmax$, e.g., $\lmax=\lambda_0 > \lambda_1 > \cdots > \lambda_K = \lmax/10^8$. 
\State Initialize $\lmin\gets\lmax$ and $k\gets 0$.
\While{$\mathcal{I}([\lmin, \lmax]) < C$}
\State $\hat{S}_\lambda(Z_{A_b}) = \{j : \Phi_{Z_{A_b}}(j) \geq \lambda\}$ for $b=1,\dots,2B$.
\State $\lmin\gets\lambda_{k+1}$ followed by $k\gets k+1$.
\EndWhile
\State $\Lambda \gets [\lmin,\lmax]$.
\State Evaluate estimated selection probability $\hat\pi_j(\lambda)=\frac{1}{2 B}\sum_{b=1}^{2 B} \1\!\big(j \in \hat{S}_\lambda(Z_{A_b})\big)$ for $j=1,\dots,p$.
\State Evaluate the integral $\mathsmaller{\int}_\Lambda f(\hat{\pi}_j(\lambda))\mu(d\lambda)$ for $j=1,\dots,p$.
\Ensure{$\efp_{Z_{1:n}}(j) = \mathcal{I}(\Lambda) / \mathsmaller{\int}_\Lambda f(\hat{\pi}_j(\lambda))\mu(d\lambda)$ for $j=1,\dots,p$.}
\end{algorithmic}
\end{algorithm}


\subsection{IPSS theory and efp scores.}\label{sup_sec:theory}


Given the estimated selection probabilities $\hat\pi_j$, an interval $\Lambda \subseteq (0,\infty)$, a probability measure $\mu$ on $\Lambda$, a function $f:[0,1]\to\mathbb{R}$, and a threshold $\tau$, \cite{ipss} define the set of features selected by IPSS by
\begin{align}\label{eq:ipss}
	\hat{S}_{\mathrm{IPSS}}(\tau) &= \Big\{j : \mathsmaller{\int}_\Lambda f(\hat{\pi}_j(\lambda))\mu(d\lambda) \geq \tau\Big\}.
\end{align}
The integral incorporates information about the selection probabilities over all of $\Lambda$, eliminating the need to select features based on individual $\lambda$ values. The function $f$ transforms the selection probabilities for better performance. Only certain choices of $f$ are known to yield valid efp scores. \cite{ipss} prove the following result (see Theorem 4.1 therein). Let $q(\lambda) = \mathrm{E}\lvert \hat{S}_\lambda(Z_{1:\nhalf})\rvert$ be the expected number of features selected by $\hat{S}_\lambda$ on half the data and, for $m\in\mathbb{N}$, define $h_m(x) = (2x - 1)^m\1(x\geq 0.5)$. For any $\Lambda$ and $\mu$, if
\begin{align}\label{eq:simultaneous}
    \max_{j\in S^c}\,\mathbb{P}\bigg(j\in\bigcap_{b=1}^{m'} \big(\hat{S}_\lambda(Z_{A_{2b-1}})\cap \hat{S}_\lambda(Z_{A_{2b}})\big)\bigg) &\leq (q(\lambda)/p)^{2m'},
\end{align}
for all $\lambda\in\Lambda$ and $m'\in\{1,\dots,m\}$, then IPSS with $f=h_m$ satisfies
\begin{align}\label{eq:ipss_bound_general}
	\mathrm{E}\lvert \hat{S}_{\mathrm{IPSS}}(\tau) \cap S^c\rvert &\leq \frac{\mathcal{I}_m(\Lambda)}{\tau}
\end{align}
for a constant $\mathcal{I}_m(\Lambda)$ whose explicit form is given in Theorem 4.1 in \cite{ipss}. Thus, setting $\efp_{Z_{1:n}}(j) = \mathcal{I}_m(\Lambda) / \int_\Lambda h_m(\hat\pi_j(\lambda)) \mu(d\lambda)$ and $\hat{S}(t)=\{j : \efp_{Z_{1:n}}(j)\leq t\}$, we have
\begin{align*}
	\hat{S}(t) &= \bigg\{j : \frac{\mathcal{I}_m(\Lambda)}{\int_\Lambda h_m(\hat\pi_j(\lambda)) \mu(d\lambda)} \leq t\bigg\}
		= \bigg\{j : \int_\Lambda h_m(\hat\pi_j(\lambda)) \mu(d\lambda) \geq \frac{\mathcal{I}_m(\Lambda)}{t}\bigg\}
		= \hat{S}_{\mathrm{IPSS}}\bigg(\frac{\mathcal{I}_m(\Lambda)}{t}\bigg)
\end{align*}
and hence
\begin{align*}
	\mathrm{E}(\mathrm{FP}(t)) &= \mathrm{E}\lvert \hat{S}(t)\cap S^c\rvert
		= \mathrm{E}\big\lvert \hat{S}_{\mathrm{IPSS}}\big(\tfrac{\mathcal{I}_m(\Lambda)}{t}\big)\cap S^c\big\rvert
		\leq t,
\end{align*}
where the inequality holds by \cref{eq:ipss_bound_general}.

The above derivation shows how valid efp scores for IPSS with $f=h_m$ are obtained under the conditions of Theorem 4.1 in \cite{ipss}. In particular, we see that by defining efp scores as above, the set $\hat{S}(t)=\{j : \efp_{Z_{1:n}}(j)\leq t\}$ is identical to $\hat{S}_{\mathrm{IPSS}}(\tau)$ when $\tau = \mathcal{I}_m(\Lambda)/t$. The main condition of the theorem, \cref{eq:simultaneous}, upper bounds the maximum probability that an unimportant feature is simultaneously selected on both halves of the data, $Z_{A_{2b-1}}$ and $Z_{A_{2b}}$, in $m'$ independent tries; see \cite{ipss} for details. The following result, part of Theorem 4.2 in \cite{ipss}, gives the form of $\mathcal{I}_3(\Lambda)$, which corresponds to the function $f=h_3$ that is used for IPSS throughout the main text.

\begin{theorem}\label{thrm:ipss}
Let $\mu$ be a probability measure on $\Lambda \subseteq (0,\infty)$, let $\tau\in(0,1]$, and define $\hat{S}_{\mathrm{IPSS}}(\tau)$ as in \cref{eq:ipss} with $f=h_3$. If \cref{eq:simultaneous} holds for all $\lambda\in\Lambda$ and $m'\in\{1,2,3\}$, then
\begin{align}\label{eq:ipss_bound}
	\mathrm{E}(\mathrm{FP}(\tau)) &\leq \frac{1}{\tau}\int_\Lambda \bigg(\frac{q(\lambda)^2}{B^2p} + \frac{3q(\lambda)^4}{Bp^3} + \frac{q(\lambda)^6}{p^5}\bigg)\mu(d\lambda),
\end{align}
where $\mathrm{E}(\mathrm{FP}(\tau)) = \mathrm{E}\lvert \hat{S}_{\mathrm{IPSS}}(\tau) \cap S^c\rvert$ is the expected number of false positives selected by IPSS.
\end{theorem}
For some intuition about the bound in \cref{eq:ipss_bound}, observe that taking $B\to\infty$ yields $\mathrm{E}(\mathrm{FP}(\tau)) \leq \tau^{-1}\int_\Lambda \big(q(\lambda)^6/p^5\big) \mu(d\lambda)$. 
In comparison, other versions of stability selection upper bound $\mathrm{E}(\mathrm{FP}(\tau))$ by $q(\lambda)^2/p$ \citep{mb,shah}, which is orders of magnitude larger than $q(\lambda)^6/p^5$ when $q(\lambda)\ll p$, as is often the case for most values of $\lambda$ in $\Lambda$. This and the contribution of $B$ in the denominators in \cref{eq:ipss_bound} largely explain the strength of the IPSS bound in \cref{thrm:ipss} relative to previous bounds, and hence the tightness of the efp scores of IPSS with $f=h_3$ relative to the efp scores of other versions of stability selection. Further theoretical and empirical comparisons between \cref{eq:ipss_bound} and other stability selection bounds are available in \cite{ipss}.


\subsection{Parameters.}\label{sup_sec:parameters}


\cref{tab:parameters} shows the default IPSS parameters used for \texttt{IPSSGB} and \texttt{IPSSRF} throughout this work (default gradient boosting and random forest parameters are in \cref{sup_sec:ipssgb} and \cref{sup_sec:ipssrf}). As noted above, our choice of function $f(x)=(2x-1)^3\1(x\geq 0.5)$ is determined by the availability and strength of the theoretical bound in \cref{thrm:ipss}. Similarly, the choice of $\nhalf$ samples used to construct the selection probabilities $\hat{\pi}_j(\lambda)$ is required for stability selection theorems (not just \cref{thrm:ipss}) to hold, and it is unclear how to adapt their proofs to accommodate other sample sizes. Thus, $f$ and $\nhalf$ are theoretically determined rather than free parameters. 

The interval $\Lambda=[\lmin,\lmax]$ is determined by setting $\lmax$ large enough that no features are selected (see, for example, \cref{step:lmax} in \cref{alg:ipss}), and setting $\lmin$ such that the integral
\begin{align*}
	\mathcal{I}(\Lambda) &= \int_\Lambda \bigg(\frac{q(\lambda)^2}{B^2p} + \frac{3q(\lambda)^4}{Bp^3} + \frac{q(\lambda)^6}{p^5}\bigg)\mu(d\lambda),
\end{align*}
in \cref{eq:ipss_bound} is equal to a fixed cutoff $C$. 
As noted in the main text, we always use $C = 0.05$, but results are largely independent of this choice (\cref{fig:sensitivity_cutoff_reg,fig:sensitivity_cutoff_class}). \cref{fig:sensitivity_delta_reg,fig:sensitivity_delta_class} show IPSS is also robust to the parameter $\delta$ that determines the measure $\mu_\delta(d\lambda) = z_\delta^{-1}\lambda^{-\delta}d\lambda$, where the normalizing constant $z_\delta$ is easily computed in closed form \citep{ipss}. The probability measure $\mu_1$ averages over $\Lambda$ on a log scale, while $\mu_0$ averages on a linear scale. 

The insignificance of $C$ and $\delta$ is unsurprising: Intuitively, the efp scores for IPSS depend primarily on $f$ and the integrand in $\mathcal{I}(\Lambda)$. The actual value of $\mathcal{I}(\Lambda)$ is much less important since the efp scores depend on the relative quantities $\mathcal{I}(\Lambda)/\int_\Lambda f(\hat\pi_j(\lambda))\mu(d\lambda)$ rather than on $\mathcal{I}(\Lambda)$ itself. Since $C$ and $\mu$ only affect the value of the bound, not the integrand, they contribute little to the actual performance of IPSS, as indicated by the sensitivity analyses in \cref{sup_sec:sensitivity}.

\begin{table*}[ht]
\centering
\begin{tabular}{lcccccc}
\toprule
\textbf{Method} & $B$ & $C$ & $f(x)$ & $\delta_\text{reg}$ & $\delta_\text{class}$ \\
\midrule
\texttt{IPSSGB} & $100$ & $0.05$ & $(2x - 1)^3\1(x \geq 0.5)$ & $1.25$ & $1$ \\ 
\texttt{IPSSRF} & $50$ & $0.05$ & $(2x - 1)^3\1(x \geq 0.5)$ & $1.25$ & $1.25$ \\
\bottomrule
\end{tabular}
\caption{\textit{Default IPSS parameters}. Both \texttt{IPSSGB} and \texttt{IPSSRF} always use $C=0.05$ and $f(x)=(2x - 1)^3\1(x \geq 0.5)$. \texttt{IPSSRF} uses $B=50$ to reduce runtimes without any noticeable difference in selection performance. The parameters $\delta_\text{reg}$ and $\delta_\text{class}$ determine the measure $\mu_\delta$ in regression and classification problems, respectively.}
\label{tab:parameters}
\end{table*}


\section{Overview of methods and implementation details}\label{sup_sec:methods}


We describe the preselection procedure used to improve feature selection (\cref{sup_sec:preselection}) and provide implementation details for each feature selection method considered in this work (\cref{sup_sec:method_list}).


\subsection{Preselection.}\label{sup_sec:preselection}


Many feature selection methods employ some form of screening, or \textit{preselection}, as an initial step in the selection process. This can be especially helpful in high dimensions for increasing power and reducing runtimes. For many of the methods in \cref{sup_sec:method_list}, we preselect features by running a randomized importance function on the full dataset three times---that is, computing $\Phi_{Z_{1:n}}$ three times---and keeping only the $p_{\text{pre}}$ features with the largest average scores across all three trials. For example, preselection for \texttt{IPSSRF} entails fitting three random forests to the full dataset and keeping the $p_\text{pre}$ features with the largest average importance scores.

For IPSS, this preselection step does not affect the theoretical control on E(FP) since $\lvert\hat{S}_{\mathrm{IPSS},\mathrm{pre}} \cap S^c\rvert = \lvert\hat{S}_{\mathrm{IPSS},\mathrm{pre}} \cap S_{\mathrm{pre}}^c\rvert$, where $\hat{S}_{\mathrm{IPSS},\mathrm{pre}}$ are the features selected by IPSS using only the preselected features, and $S_{\mathrm{pre}}^c$ are the preselected features in $S^c$. Of course, preselection risks discarding important features, potentially increasing the number of false negatives. In practice, however, we find that preselection actually helps IPSS, stability selection, and model-X knockoffs identify more true positives while still controlling false discoveries. This is because preselection gets rid of many noisy features, making it easier for these methods to detect the true signal.

In \cref{sup_sec:method_list}, we describe the preselection parameters for each method, which were determined by extensive testing on simulated data. We do not apply preselection when implementing \texttt{Boruta}, \texttt{RFEGB}, \texttt{RFHT}, \texttt{Vita}, and \texttt{VSURF} since these methods include their own internal screening steps.


\subsection{Methods.}\label{sup_sec:method_list}


We describe how each method in \cref{tab:methods} is implemented in this work.

\begin{table*}[ht]
    \centering
    \begin{tabular}{lcccc}
        \toprule
        \textbf{Method} & \textbf{Package} & \textbf{Error control} & \textbf{Base method} & \textbf{Non-default settings}  \\
        \midrule
        \texttt{IPSSGB} & \textit{ipss} (Python) & \darkgreen{\ding{51}} & Boosting &  --- \\\hline
        
        \texttt{IPSSRF} & \textit{ipss} (Python) & \darkgreen{\ding{51}} & Random forest &  --- \\\hline
        
        \texttt{IPSSL1} & \textit{ipss} (Python) & \darkgreen{\ding{51}} & Lasso & --- \\\hline
        
        \texttt{KOGLM} & \textit{knockoff} (R) & \darkgreen{\ding{51}} & GLM & --- \\\hline
        
        \texttt{KOL1} & \textit{knockoff} (R) & \darkgreen{\ding{51}} & Lasso & --- \\\hline
        
        \texttt{KORF} & \textit{knockoff} (R) & \darkgreen{\ding{51}} & Random forest & --- \\\hline
        
        \texttt{DeepPINK} & \textit{knockpy} (Python) & \darkgreen{\ding{51}} & Neural network & --- \\\hline
        
		\multirow{2}{*}{\texttt{SSBoost}} & \textit{XGBoost} (Python) & \multirow{2}{*}{\darkgreen{\ding{51}}} & \multirow{2}{*}{Boosting} & $\texttt{assumption} = r\mathrm{-concave}$ \\
    & with \textit{stabs} (R) &  &  & $\tau = 0.75$ \\\hline
    
        \multirow{2}{*}{\texttt{KOBT}} & \multirow{2}{*}{\textit{KOBT} (R)} & \multirow{2}{*}{\darkgreen{\ding{51}}} & \multirow{2}{*}{Boosting} & $\texttt{num}=100$, $\texttt{bound}=200$, \\ &&&&$\texttt{type}=$ shrink \\\hline
        
        \texttt{RFHT} & \textit{rfvimptest} (R) & \darkgreen{\ding{51}} & Random forest & --- \\\hline
        
        \texttt{Boruta} & \textit{Boruta} (R) & \red{\ding{55}} & Random forest &  --- \\\hline
        
        \texttt{RFEGB} & \textit{scikit-learn} (Python) & \red{\ding{55}} & Boosting & --- \\\hline
        
        \texttt{Vita} & \textit{vita} (R) & \red{\ding{55}} & Random forest & $p$-value threshold $= 0$ \\\hline
        
        \texttt{VSURF} & \textit{VSURF} (R) & \red{\ding{55}} & Random forest &\texttt{VSURF\_pred} \\
        \bottomrule
    \end{tabular}
    \caption{\textit{Feature selection methods}. Software packages are listed with the language used to implement them in parentheses. Details about each method, including descriptions of their non-default settings, are in Sections~\ref{sup_sec:ipssgb}--\ref{sup_sec:methods_no_control}. For methods with no non-default settings, we use the default settings in their respective packages.}
\label{tab:methods}
\end{table*}


\subsubsection{\texttt{IPSSGB}.}\label{sup_sec:ipssgb}


The IPSS-related parameters used to implement \texttt{IPSSGB} are in \cref{tab:parameters}. For preselection, we use gradient boosting as the baseline selection algorithm and set $p_\text{pre}=100$. We implement gradient boosting using \texttt{XGBoost} \citep{xgboost}. All \texttt{XGBoost} parameters are set to their default values except for two changes: The proportion of features considered when splitting each node (called \texttt{colsample\_bynode} in \texttt{XGBoost} and often $\mathtt{mtry}$ elsewhere) is changed from $1$ to $1/3$, and the maximum depth of each tree (\texttt{max\_depth}) is changed from $6$ to $1$, making each tree a stump. The latter change significantly improved the performance of \texttt{IPSSGB}, both in terms of speed and feature selection.


\subsubsection{\texttt{IPSSRF}.}\label{sup_sec:ipssrf}


The IPSS-related parameters used to implement \texttt{IPSSRF} are in \cref{tab:parameters}. For preselection, we use random forests as the baseline selection algorithm and set $p_\text{pre}=100$. We implement random forests using \texttt{scikit-learn} \citep{sklearn}. All random forest parameters are set to their default values except for two changes: The proportion of features considered when splitting each node (called \texttt{max\_features} in \texttt{scikit-learn}) is changed from $1$ to $1/10$, and the number of trees used to build each random forest (\texttt{n\_estimators}) is changed from $100$ to $50$. These changes improved the efficiency of \texttt{IPSSRF} without sacrificing its feature selection performance.


\subsubsection{\texttt{IPSSL1}.}\label{sup_sec:ipssl1}


As discussed in the main text, \texttt{IPSSL1} is a parametric version of IPSS based on $\ell^1$-regularization \citep{ipss}. For regression, the baseline algorithm is lasso \citep{lasso}, and for classification, it is $\ell^1$-regularized logistic regression \citep{logistic_regression}. All parameters are set to their default values in the \texttt{ipss} Python package: \href{https://pypi.org/project/ipss/}{https://pypi.org/project/ipss/}. 


\subsubsection{Model-X knockoffs (\texttt{KOGLM}, \texttt{KOL1}, \texttt{KORF}, \texttt{DeepPINK}, \texttt{KOBT}).}\label{sup_sec:knockoffs}


Model-X knockoffs work by first constructing \textit{knockoffs} $\tilde{X} = (\tilde{X}_1,\dots,\tilde{X}_p)$ of the original features $X = (X_1,\dots,X_p)$. By definition, $\tilde{X}$ must satisfy (i) the joint distribution of $(X,\tilde{X})$ is invariant under pairwise exchanges of the original features $X_j$ and their corresponding knockoffs $\tilde{X}_j$, and (ii) $\tilde{X}$ is conditionally independent of the $Y$ given $X$ (see Definition 2 in \cite{knockoffs}). Once knockoffs are constructed, feature importance scores---called \textit{feature statistics} in the knockoffs literature---are computed for all of the original and knockoff features and subsequently used to select original features in a way that controls the FDR. Like IPSS, any feature importance function can be used. 

As noted in the main text, model-X knockoffs requires knowledge of the joint distribution of $X$. In our multivariate Gaussian simulations in \cref{sec:simulation_studies}, all of the model-X knockoffs methods are implemented using the true, known joint distribution. In the remaining examples, where the joint distribution of $X$ is not known, we use the default methods for constructing approximate model-X knockoffs (for example, second-order Gaussian knockoffs in the \texttt{knockoff} R package).

\texttt{KOGLM}, \texttt{KOL1}, and \texttt{KORF} are all implemented using the R package \texttt{knockoff}. \texttt{KOGLM} uses feature importance scores from a generalized linear model (GLM); we find it performs best when using random forests for preselection with $p_\text{pre}=200$. \texttt{KOL1} uses feature importance scores from lasso (regression) or $\ell^1$-regularized logistic regression (classification). Like \texttt{IPSSL1}, we use lasso (or logistic regression) for preselection, with $p_\text{pre}=200$. \texttt{KORF} uses feature importance scores from random forests; like \texttt{KOGLM}, we find it performs best when using random forests for preselection with $p_\text{pre}=200$.

\texttt{DeepPINK} uses deep neural networks to construct feature importance scores. We implement it using the Python package \texttt{knockopy} and random forests for preselection with $p_\text{pre}=100$.

\texttt{KOBT} is implemented with the R package \texttt{KOBT}. It uses boosted tree models to construct importance scores. We tested \texttt{KOBT} numerous times on simulated data with many different tuning and preselection parameters (including no preselection), but found that \texttt{KOBT} consistently and dramatically exceeded its target FDR. For this reason, we omit its results from all plots and tables in this work.

Finally, we tested all of the model-X knockoffs methods without using preselection. In these cases, power was often extremely low and the runtimes were much longer.


\subsubsection{\texttt{SSBoost}.}\label{sup_sec:ssboost}


The closest method to \texttt{IPSSGB} in terms of its underlying approach is that of \cite{ssboost}, referred to here as \texttt{SSBoost}. Unlike \texttt{IPSSGB}---which uses importance scores from gradient boosting---\texttt{SSBoost} applies stability selection to choose the number of features used per boosting run. Furthermore, \texttt{IPSSGB} uses IPSS to construct efp scores, whereas \texttt{SSBoost} uses a version of stability selection introduced by \cite{shah}. This is perhaps the most significant difference since the efp scores for IPSS have much tighter bounds than those for other forms of stability selection \citep{ipss}. From a practical standpoint, this causes other versions of stability selection to identify fewer important features than IPSS.

\cite{ssboost} provide code for \texttt{SSBoost} that combines the R packages \texttt{mboost} \citep{mboost} and \texttt{stabs} \citep{stabs} in the form of a worked example, but the \texttt{mboost} implementation of boosting was prohibitively slow in the dimensions we consider. By adapting their code to use \texttt{XGBoost} in place of \texttt{mboost} for boosting and by preselecting features using gradient boosting with $p_\text{pre}=150$, we were able to reduce \texttt{SSBoost} runtimes considerably with no apparent change in results. The \texttt{XGBoost} parameters used for \texttt{SSBoost} are the same as those used for \texttt{IPSSGB} (\cref{sup_sec:ipssgb}). For the stability selection part of \texttt{SSBoost}, we use the default parameters in \texttt{stabs}, and the selection threshold is set to $\tau=0.75$, which is the middle of the interval $(0.6,0.9)$ recommended by \cite{mb}.


\subsubsection{\texttt{RFHT}.}\label{sup_sec:rfht}


We tested \texttt{RFHT} \citep{pitt}, which achieves theoretical error control by using random forests for hypothesis testing. However, one test run with default parameters on simulated data with 500 samples, 500 features, and 20 true features took over 51 minutes, returning 15 true positives and 43 false positives. By contrast, \texttt{IPSSGB} with a target FDR of $0.2$ took 11 seconds and returned 10 true positives and 0 false positives on the same data. This method is omitted from our studies because its performance does not appear to justify its excessive runtime. 


\subsubsection{Methods without false discovery control (\texttt{Boruta}, \texttt{RFEGB}, \texttt{Vita}, \texttt{VSURF}).}\label{sup_sec:methods_no_control}


\texttt{Boruta}, \texttt{Vita}, and \texttt{VSURF} are implemented using R packages of the same names. \texttt{Boruta} is run with default parameters. For \texttt{Vita}, we set the $p$-value threshold to $0$. For \texttt{VSURF}, we use the function \texttt{VSURF\_pred} rather than \texttt{VSURF\_interp} to select the final set of features \citep{vsurf}. Both of these choices favor sparsity, which aligns well with our simulation designs. As noted in the main text, \texttt{VSURF} is too computationally expensive to include in our $p=2000$ and $5000$ simulation studies; see also \cref{tab:runtimes_reg}. 

We implement \texttt{RFEGB} by combining \texttt{XGBoost} and \texttt{scikit-learn}. On simulated Gaussian data with $n=250$ and $p=500$, one run of \texttt{RFEGB} took over 12 minutes when removing five features per iteration (the default is one feature removed per iteration, which takes approximately 5 times as long). For comparison, \texttt{IPSSGB} ran in 5 seconds on the same data and had far fewer false positives and similar power. Due to its high computational cost, poor performance, and many tuning parameters (which is true of recursive feature elimination in general), \texttt{RFEGB} is largely omitted from this work.


\section{Simulation results and details}\label{sup_sec:simulations}


We present additional simulation results (\cref{sup_sec:simulation_results}) and RNA-seq simulation details (\cref{sup_sec:simulation_details}).


\subsection{Additional simulation results.}\label{sup_sec:simulation_results}

\begin{figure*}[ht]
\includegraphics[height=.375\textheight, width=\textwidth]{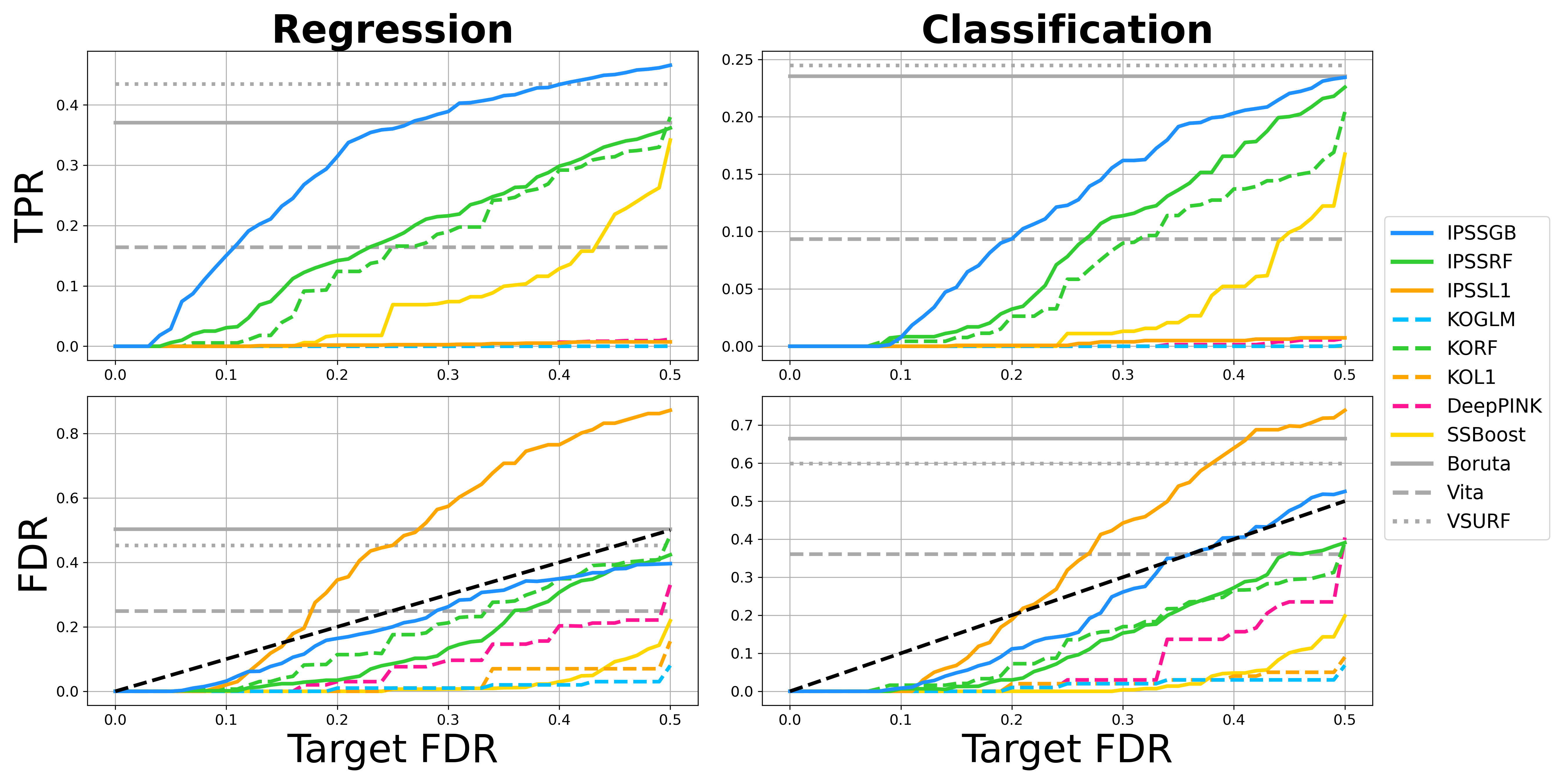}
\caption{\textit{Gaussian simulation results ($n=250$)}. See \cref{fig:toeplitz_0.5_n500} for details.}
\label{fig:toeplitz_0.5_n250}
\end{figure*}
\begin{figure*}[ht]
\includegraphics[height=.325\textheight, width=\textwidth]{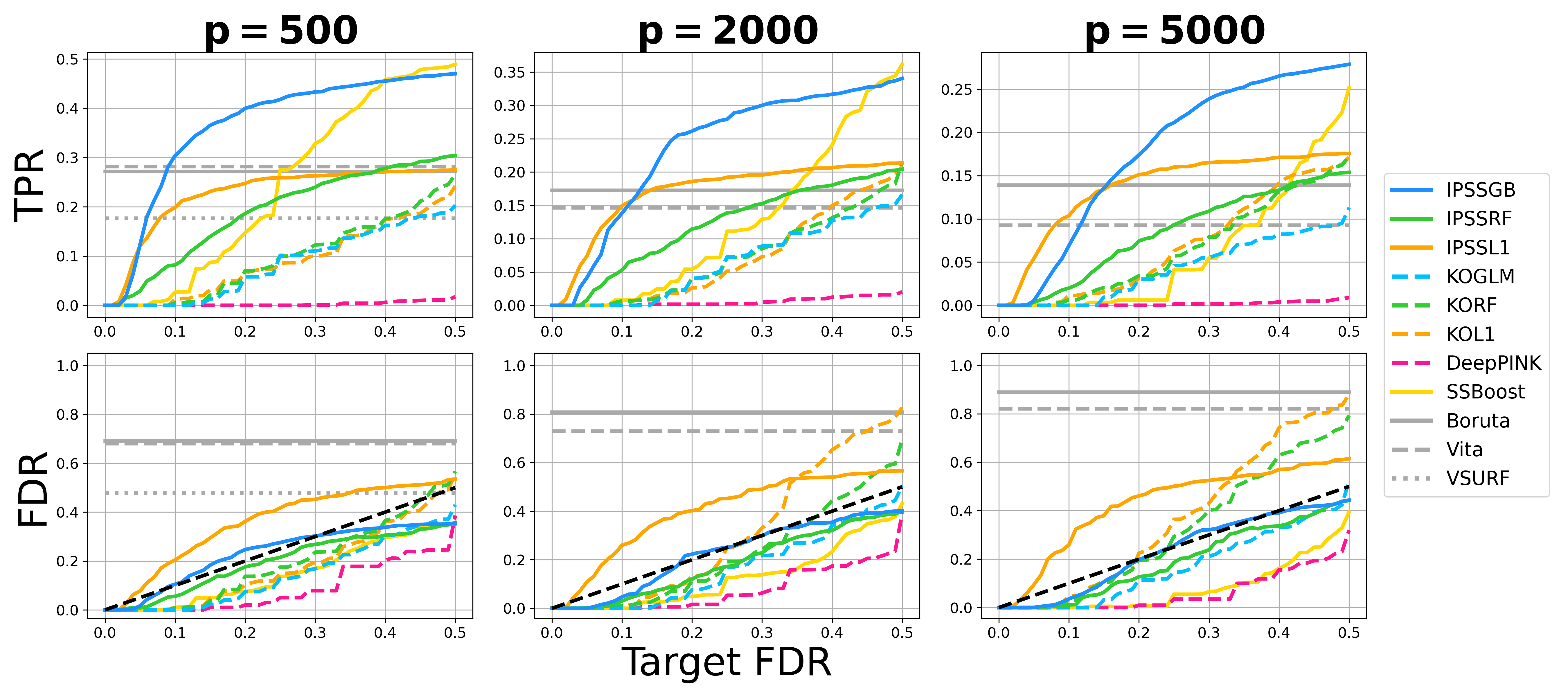}
\caption{\textit{RNA-seq simulation results (classification)}. See \cref{fig:oc_reg} for details.}
\label{fig:oc_class}
\end{figure*}

\cref{fig:toeplitz_0.5_n250} shows the $n=250$ multivariate Gaussian simulation results, described in \cref{sec:gaussian_simulations}. \cref{fig:oc_class} shows the $p=500$, $2000$, and $5000$ RNA-seq simulation results for classification, described in \cref{sec:rnaseq_simulations}. \cref{tab:runtimes_reg,tab:runtimes_class} show the average runtimes of each method in each simulation experiment. 

\begin{table}[ht]
\centering
\begin{tabular}{lccccc}
\toprule
\textbf{Method} & MG($n=250$) & MG($n=500$) & RNA($p=500$) & RNA($p=2000$) & RNA($p=5000$) \\
\midrule
\texttt{IPSSGB} & 4.4 (0.11) & 6.3 (0.01) & 6.7 (0.23) & 9.0 (0.43) & 14.6 (0.45) \\
\texttt{IPSSRF} & 4.5 (0.05) & 7.5 (0.10) & 7.2 (0.14) & 10.8 (0.29) & 19.6 (0.63) \\
\texttt{IPSSL1} & 1.6 (0.23) & 1.0 (0.05) & 4.0 (0.91) & 3.1 (0.66) & 6.4 (0.76) \\
\texttt{KOGLM} & 10.8 (0.92) & 12.4 (0.91) & 10.0 (1.92) & 12.6 (1.50) & 19.2 (1.66) \\
\texttt{KOL1} & 11.2 (0.88) & 12.0 (1.04) & 7.5 (1.27) & 6.7 (0.93) & 7.2 (1.16) \\
\texttt{KORF} & 12.8 (1.16) & 14.3 (1.01) & 4.2 (0.50) & 7.4 (0.22) & 13.9 (0.50) \\
\texttt{DeepPINK} & 5.0 (3.55) & 6.0 (2.79) & 5.9 (4.17) & 8.8 (0.21) & 15.4 (0.42) \\
\texttt{SSBoost} & 4.1 (0.04) & 6.7 (0.03) & 6.7 (0.04) & 8.1 (0.15) & 13.7 (0.08) \\
\texttt{Boruta} & 42.2 (3.09) & 120.7 (7.93) & 42.0 (5.96) & 57.0 (6.51) & 84.5 (6.66) \\
\texttt{Vita} & 9.5 (0.19) & 24.2 (0.40) & 20.6 (0.47) & 83.5 (4.09) & 195.3 (4.34) \\
\texttt{VSURF} & 196.8 (8.16) & 548.6 (23.68) & 286.7 (72.36) & --- & ---  \\
\texttt{RFHT} & --- & 3087* & --- & --- & ---  \\
\texttt{RFEGB} & 749* & --- & --- & --- & --- \\
\bottomrule
\end{tabular}
\caption{Average runtimes (in seconds) over 100 trials for each regression experiment. MG stands for Multivariate Gaussian. Standard deviations are in parentheses. Recall that $p=500$ in the MG experiments and $n=500$ in the RNA experiments, and that \texttt{VSURF} was too computationally expensive to include when $p=2000$ and $5000$. *As discussed in \cref{sup_sec:rfht,sup_sec:methods_no_control}, \texttt{RFHT} and \texttt{RFEGB} are largely omitted due to their poor selection performance and excessive runtimes in initial tests (shown in the table). Hence, their remaining entries are blank.}
\label{tab:runtimes_reg}
\end{table}

\begin{table}[ht]
\centering
\begin{tabular}{lccccc}
\toprule
\textbf{Method} & MG($n=250$) & MG($n=500$) & RNA($p=500$) & RNA($p=2000$) & RNA($p=5000$) \\
\midrule
\texttt{IPSSGB} & 4.3 (0.04) & 6.5 (0.11) & 6.6 (0.07) & 7.5 (0.20) & 14.3 (0.42) \\
\texttt{IPSSRF} & 4.2 (0.02) & 7.0 (0.05) & 6.8 (0.10) & 9.9 (0.22) & 17.7 (0.49) \\
\texttt{IPSSL1} & 5.9 (0.15) & 6.0 (0.22) & 8.4 (0.86) & 9.3 (1.17) & 13.1 (1.25) \\
\texttt{KOGLM} & 10.6 (0.10) & 12.0 (0.14) & 9.3 (1.57) & 11.8 (1.67) & 17.4 (1.64) \\
\texttt{KOL1} & 10.7 (0.12) & 11.9 (0.21) & 7.2 (1.13) & 6.0 (0.83) & 6.3 (0.80) \\
\texttt{KORF} & 10.1 (0.95) & 12.1 (0.17) & 3.1 (0.10) & 5.9 (0.18) & 11.3 (0.38) \\
\texttt{DeepPINK} & 4.7 (0.29) & 5.7 (0.28) & 5.2 (0.10) & 8.0 (0.24) & 13.3 (0.43) \\
\texttt{SSBoost} & 4.0 (0.01) & 6.6 (0.04) & 6.7 (0.02) & 8.2 (0.04) & 13.8 (0.11) \\
\texttt{Boruta} & 27.8 (1.81) & 76.6 (6.29) & 25.4 (3.91) & 37.4 (4.51) & 58.6 (3.99) \\
\texttt{Vita} & 4.8 (0.06) & 10.7 (0.06) & 9.2 (0.16) & 35.3 (0.72) & 89.1 (4.59) \\
\texttt{VSURF} & 78.1 (4.14) & 190.9 (8.93) & 77.0 (28.08) & --- & --- \\
\bottomrule
\end{tabular}
\caption{Average runtimes (in seconds) over 100 trials for each classification experiment. MG stands for Multivariate Gaussian. Standard deviations are in parentheses. Recall that $p=500$ in the MG experiments and $n=500$ in the RNA experiments. \texttt{VSURF} was too computationally expensive to include when $p=2000$ and $5000$.}
\label{tab:runtimes_class}
\end{table}

\clearpage


\subsection{RNA-seq simulation details.}\label{sup_sec:simulation_details}


\cref{alg:simulation} describes the data generating procedure for the RNA-seq simulation studies (\cref{sec:rnaseq_simulations}), which is also depicted in \cref{fig:simulation}. In all steps, ``randomly select" means select a parameter uniformly at random from its domain, which are shown in \cref{tab:simulation_parameters}. The randomized function $f_\theta:\mathbb{R}\to [-1,1]$ that links the features to the response is
\begin{align}
\label{eq:f_theta}
	f_\theta(x) &=
	\begin{cases}
		\frac{\delta_1}{2}\left(1 + \tanh(\alpha(\delta_2 x - \beta))\right) & \text{with probability}\ 1/2, \\
		\delta_1\exp(-\gamma x^2) & \text{with probability}\ 1/2,
	\end{cases}
\end{align}
where each component of $\theta = (\alpha, \beta, \gamma, \delta_1, \delta_2)$ is drawn uniformly at random prior to each trial according to \cref{tab:simulation_parameters}. The values of $\alpha$ and $\gamma$ determine steepness of the curves, $\beta$ shifts $\tanh$ horizontally, and $\delta_1$ and $\delta_2$ reflect the functions about the horizontal and vertical axes, respectively. \cref{fig:functions} shows five realizations of $f_\theta$, illustrating the many ways it can influence the response, $Y$. For example, if the function in the left-most panel of \cref{fig:functions} is applied to the genes in a given partition of $S$, then those genes will only significantly affect $Y$ if their collective expression level is positive, while collective expression levels less than $-1$ will have virtually no effect on $Y$.

\begin{table*}[!ht]
\centering
\begin{tabular}{cccccc}
\toprule
\textbf{Parameter} & $\alpha$ & $\beta$ & $\gamma$ & $\delta_1$ & $\delta_2$ \\
\midrule
\textbf{Range} & $(0.5,1.5)$ & $(-1,1)$ & $(1,3)$ & $\{-1,1\}$ & $\{-1,1\}$ \\
\bottomrule
\end{tabular}
\caption{\textit{Simulation parameters}. Parameters are drawn uniformly at random from their corresponding ranges prior to each simulation trial.}
\label{tab:simulation_parameters}
\end{table*}

\begin{algorithm}[!ht]
\caption{Data generation for simulation study (one trial)}\label{alg:simulation}
\begin{algorithmic}[1]
\Require{RNA-seq data $X_{\mathrm{full}}\in\mathbb{R}^{596\times 6426}$, number of samples $n$, number of features $p$, number of true features $\lvert S\rvert$, signal-to-noise ratio SNR, function parameter domain $\Theta$.}
\State Randomly select $n$ rows and $p$ columns of $X_{\mathrm{full}}$. Denote the resulting matrix by $X\in\mathbb{R}^{n\times p}$.
\State Standardize the columns of $X$ to have mean 0 and variance 1.
\State Randomly select $\lvert S\rvert$ true features $S\subseteq\{1,\dots,p\}$.
\State Randomly select $G\in\{\lfloor \lvert S\rvert/2\rfloor,\dots,\lvert S\rvert\}$. Partition $S$ into $G$ disjoint groups, $S = \bigsqcup_{g=1}^G S_g$.
\State Initialize the signal, $\eta \gets (0,\ldots,0)^\mathtt{T} \in \mathbb{R}^n$. 
\For{$g = 1,\dots,G$}
\State $\xi_g \gets \sum_{j\in S_g} X_j$ where $X_j\in\mathbb{R}^n$ is the $j$th column of $X$.
\State Standardize $\xi_g$ to have mean 0 and variance 1. 
\State Randomly select a function parameter $\theta\in\Theta$.
\State $\eta \gets \eta + f_\theta(\xi_g)$ with $f_\theta$ applied to $\xi_g\in\mathbb{R}^n$ componentwise.
\EndFor
\State For regression: Draw $\epsilon_i\sim\mathcal{N}(0,\sigma^2)$ with $\sigma^2 = \sum_{i=1}^n \eta_i^2 / (n\,\mathrm{SNR})$ and set $y_i\gets \eta_i + \epsilon_i$.
\State For classification: Draw $u\sim\mathrm{Uniform}(1,3)$, then $y_i\sim\mathrm{Bernoulli}(\pi_i)$ where $\pi_i = 1/(1+\exp(-u \eta_i))$.  
\Ensure{Features $X\in\mathbb{R}^{n\times p}$, responses $y\in\mathbb{R}^n$, and important features $S\subseteq \{1,\ldots,p\}$.}
\end{algorithmic}
\end{algorithm}

\begin{figure}[ht]
\centering
\resizebox{\textwidth}{!}{
\begin{tikzpicture}[
    every node/.style={align=center},
    every node/.append style={font=\Large},
    arrow/.style={-Stealth, thick}
]

\node (Xfull) {$X_{\mathrm{full}}$};

\node[right=1cm of Xfull] (X) {$X$};

\draw[arrow] (Xfull) -- (X);

\node[right=1cm of X] (XS) {$X_S$};
\node[below=0.5cm of X] (XN) {$X_{S^c}$};

\draw[arrow] (X) -- (XS);
\draw[arrow] (X) -- (XN);

\node[right=2.5cm of XS, yshift=1cm] (XS1) {$X_{S_1}$};
\node[right=2.5cm of XS] (XS2) {$X_{S_2}$};
\node[right=2.5cm of XS, yshift=-1cm] (XS3) {$X_{S_3}$};

\draw[arrow] (XS) -- (XS1);
\draw[arrow] (XS) -- (XS2);
\draw[arrow] (XS) -- (XS3);

\node[right=1cm of XS1] (xi1) {$\xi_1$};
\node[right=1cm of XS2] (xi2) {$\xi_2$};
\node[right=1cm of XS3] (xi3) {$\xi_3$};

\draw[arrow] (XS1) -- (xi1);
\draw[arrow] (XS2) -- (xi2);
\draw[arrow] (XS3) -- (xi3);

\node[right=2.5cm of xi2] (y) {$Y = \sum_{g=1}^3 f_{\theta_g}(\xi_g) + \epsilon$};

\draw[arrow] (xi1) -- (y) node[midway, above] {$f_{\theta_1}$};
\draw[arrow] (xi2) -- (y) node[midway, above] {$f_{\theta_2}$};
\draw[arrow] (xi3) -- (y) node[midway, above] {$f_{\theta_3}$};

\end{tikzpicture}
}
\caption{\textit{Simulation diagram}. Rows and columns are randomly selected from the full RNA-seq dataset to create a matrix $X$ whose columns are split into important features, $X_S$, and unimportant features, $X_{S^c}$. The columns of $X_S$ are further partitioned into $G$ groups; the above figure shows $G=3$. The features in each group are summed to obtain $\xi_g$, and a different realization $f_{\theta_g}$ of $f_\theta$ is applied to $\xi_g$ for each $g$. For regression, response is the sum of the group-specific signals, $f_{\theta_g}(\xi_g)$, plus noise.}
\label{fig:simulation}
\end{figure} 

\vspace*{2em}

\ifthenelse{\boolean{showfigures}}{
\begin{figure}[H]
\centering
\includegraphics[height=.15\textheight, width=\textwidth]{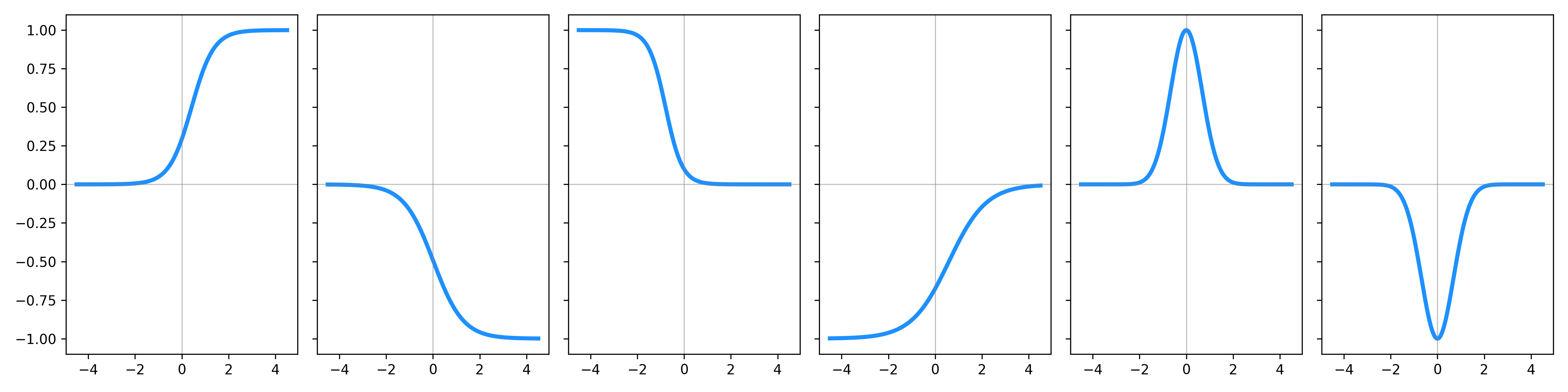}
\caption{\textit{Some realizations of the randomized function $f_\theta$}.}
\label{fig:functions}
\end{figure}
}{}

\vspace*{2em}

\ifthenelse{\boolean{showfigures}}{
\begin{figure}[H]
\includegraphics[height=.15\textheight, width=\textwidth]{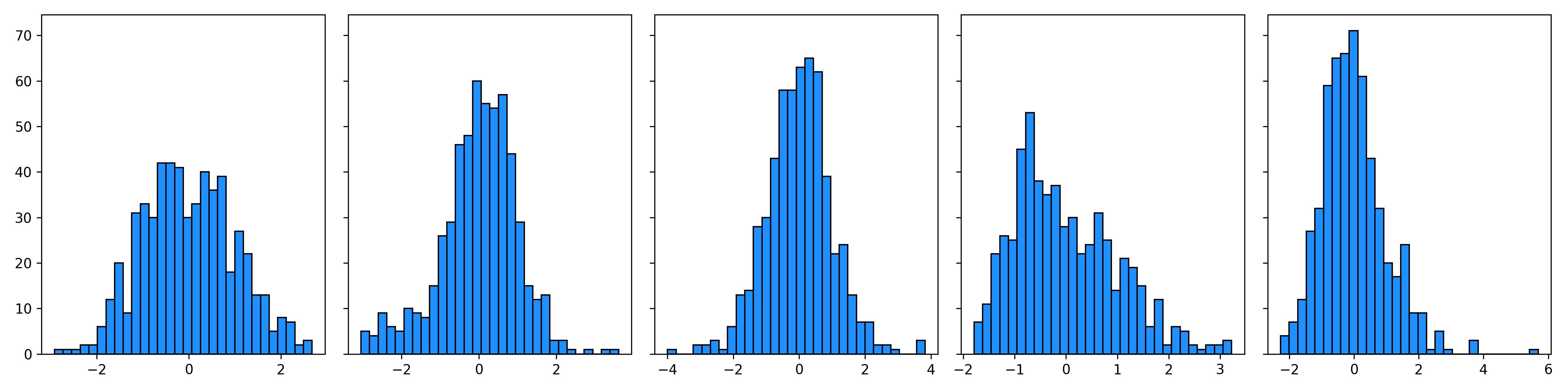}%
\caption{\textit{Distributions of five randomly selected genes from the RNA-seq dataset}. Features in the ovarian cancer RNA-seq dataset follow a variety of empirical marginal distributions. For example, the standardized empirical distributions of the five randomly selected genes above, from
left to right, are relatively flat, skewed left, approximately Gaussian, skewed right, and contain outliers. Furthermore, the genes exhibit complex correlation structures, with maximum and average absolute pairwise correlations of approximately 0.95 and 0.17 after standardization, respectively.}
\label{fig:rnaseq_features}
\end{figure}
}{}

\clearpage


\section{Cancer study results}\label{sup_sec:cancer}


We describe our ovarian cancer and glioma studies (\cref{sup_sec:ovarian,sup_sec:glioma}) and present our literature search and cross-validation results (\cref{sup_sec:lit_search,sup_sec:cv_study}).


\subsection{Ovarian cancer.}\label{sup_sec:ovarian}


For the ovarian cancer cohort, we study the following feature and response combinations (the number of samples and features are shown in parentheses): (i) miRNAs and prognosis ($n=442$, $p=585$), (ii) miRNAs and tumor purity ($n=451$, $p=585$), (iii) miRNAs and miR-150 ($n=453$, $p=584$), (iv) genes and prognosis ($n=549$, $p=6426$), and (v) genes and AKT2 ($n=569$, $p=6425$). In all cases, missing values were removed. Gene expression levels are measured by RNA-seq. We chose miR-150 and the gene AKT2 as responses in studies (iii) and (v), respectively, because literature searches indicated that both are highly related to ovarian cancer. Literature search results are only reported for clinical responses (prognosis and tumor purity).


\subsection{Glioma.}\label{sup_sec:glioma}


For the glioma cohort, we study the following feature and response combinations (the number of samples and features in each dataset are shown in parentheses): (i) miRNAs and prognosis ($n=477$, $p=787$), (ii) miRNAs and miR-155 ($n=512$, $p=786$), (iii) genes and prognosis ($n=625$, $p=10,058$), and (iv) genes and FOXM1 ($n=669$, $p=10,057$). In all cases, missing values were removed. Gene expression levels are measured by RNA-seq. The original glioma RNA-seq dataset contains over $20,000$ genes; we only study the roughly $10,000$ genes in the top 50th percentile of average gene expression (discarding lowly expressed genes is common practice). We chose miR-155 and the gene FOXM1 as responses in studies (ii) and (iv), respectively, because literature searches indicated that both are highly related to glioma. Literature search results are only reported when the response is prognosis.


\subsection{Literature search.}\label{sup_sec:lit_search}


In each of the forthcoming studies, we perform literature searches to validate our findings. For the microRNA (miRNA) and ovarian cancer prognosis study, we briefly summarize literature supporting each miRNA selected by at least one feature selection method. Providing such summaries for every study is beyond the scope of this work. Thus, to roughly quantify the relevance of selected features in subsequent studies, we searched the feature name, cancer type, and the word ``prognosis" in Europe PMC, an open-access database containing millions of life sciences publications \citep{europePMC}. We then report the total number of citations among all articles returned by the search. For example, in \cref{tab:ovarian_mirna_status}, searching ``miR-93" + ``ovarian cancer" + ``prognosis" in Europe PMC returned 43,996 citations. For each method, we also include the number of features it selected that had above a certain number of citations, below a certain number of citations, and the total number of features it selected below a certain target FDR. We emphasize that these metrics are less important than the papers themselves. Citation counts favor older publications, and our search criterion does not guarantee relevance to the specific problem at hand (though the summaries below suggest that at least some results are meaningful).

Below, we briefly summarize literature relating miRNAs in \cref{tab:ovarian_mirna_status} to ovarian cancer. Additional details are available in the associated references.

\textit{miR-1-2}. \cite{mir1-2} found that miR-1-2 is differentially expressed between cancerous and non-cancerous ovarian cancer cells, but we found no literature linking miR-1-2 to prognosis.

\textit{miR-30d}. \cite{mir30d_1} found that miR-30d suppresses ovarian cancer progression by reducing the levels of Snail, a protein involved in making cancer cells more invasive. They concluded that miR-30d could be used as a treatment for ovarian cancer. \cite{mir30d_2} found that miR-30d is associated with ``significantly better disease-free or overall survival" in ovarian cancer patients.

\textit{miR-93}. \cite{mir93_1} found that miR-93 is significantly upregulated in ovarian cancer cells that are resistant to the chemotherapy drug cisplatin. They also found that miR-93 targets the tumor suppressor gene PTEN and plays a role in the AKT signaling pathway. They concluded that further study of miR-96 may yield therapeutic strategies for overcoming cisplatin-resistant ovarian cancer cells. \cite{mir93_2} found that miR-93 is a potential biomarker of ovarian cancer.

\textit{miR-96}. \cite{mir96_1} found that overexpression of miR-96 promotes cell proliferation and migration in ovarian cancer cells. They conclude that targeting miR-96 is a potentially promising strategy for treating ovarian cancer. They also report that miR-96 inhibits phosphorylation of AKT, a gene identified by IPSSGB as being relevant to ovarian cancer prognosis. \cite{mir96_2} found that individuals with low-levels of miR-96 ``suffered more advanced tumor staging and a worse overall survival" and also identified miR-96 as a potential therapeutic target.

\textit{miR-150}. \cite{mir150_1} found significant associations between miR-150 downregulation and ``aggressive clinicopathological features" in ovarian cancer patients, as well as reduced overall and progression-free survival. They also identified miR-150 expression as a prognostic biomarker in ovarian cancer. \cite{mir150_2} found that downregulation of miR-150 is associated with resistance to paclitaxel, a chemotherapy drug used to treat ovarian cancer. They also report that treatment with pre-miR-150 resensitized cancer cells to paclitaxel, making the drug more effective.

\textit{miR-342}. \cite{mir342} found that miR-342 inhibits the proliferation, invasion, and migration of ovarian cancer cells, and promotes the death of these cells. The study also showed that miR-342 decreases the expression of key proteins involved in the Wnt/$\beta$-catenin signaling pathway, which may explain its effects on reducing ovarian cancer cell viability and growth.


\textit{miR-1270}. \cite{mir1270} found that miR-1270 plays a role in sensitivity to the chemotherapy drug cisplatin.

\textit{miR-1301}. \cite{mir1301} found that targeting miR-1301
can inhibit the proliferation of cells that are resistant to the chemotherapy drug cisplatin, thus reducing the occurrence and development of drug-resistant ovarian cancer.

\begin{table}[H]
\centering
\begin{tabular}{lccccccccc}
\toprule
\textbf{miRNA} & \textbf{Citations} & \texttt{IPSSGB} & \texttt{IPSSRF} & \texttt{IPSSL1} & \texttt{KOGLM} & \texttt{KORF} & \texttt{KOL1} & \texttt{DeepPINK} & \texttt{SSBoost} \\
\midrule
miR-93 & 43996 & 0.35 & 0.11 & -- & -- & -- & -- & -- & -- \\
miR-148a & 42177 & -- & 0.39 & -- & -- & -- & -- & -- & -- \\
miR-150 & 41195 & 0.23 & 0.39 & -- & -- & -- & -- & -- & -- \\
miR-96 & 23010 & 0.23 & -- & 0.47 & -- & -- & -- & -- & -- \\
miR-342 & 20291 & -- & 0.39 & 0.23 & -- & -- & -- & -- & -- \\
miR-30d & 19267 & 0.23 & 0.33 & -- & -- & -- & -- & -- & -- \\
miR-301b & 3224 & 0.35 & -- & -- & -- & -- & -- & -- & -- \\
miR-1270 & 2543 & 0.28 & 0.11 & 0.21 & -- & -- & -- & -- & -- \\
miR-1301 & 1390 & 0.35 & -- & -- & -- & -- & -- & -- & -- \\
miR-1-2 & 1220 & 0.35 & -- & 0.21 & -- & -- & -- & -- & -- \\
\midrule
$\geq$ 1000 & -- & 8 & 6 & 4 & 0 & 0 & 0 & 0 & 0 \\
$<$ 1000 & -- & 0 & 0 & 0 & 0 & 0 & 0 & 0 & 0 \\
Total & -- & 8 & 6 & 4 & 0 & 0 & 0 & 0 & 0 \\
\bottomrule
\end{tabular}
\caption{\textit{MicroRNAs and prognosis (ovarian cancer).} MiRNAs are ordered by citation count. A missing $q$-value indicates the miRNA was assigned a $q$-value of less than 0.5 by the corresponding method. The bottom rows report, for each method, the number of selected features with over 1000 citations, under 1000 citations, and the total number selected at the maximum target FDR of 0.5.}
\label{tab:ovarian_mirna_status}
\end{table}

\begin{table}[H]
\centering
\begin{tabular}{lccccccccc}
\toprule
\textbf{miRNA} & \textbf{Citations} & \texttt{IPSSGB} & \texttt{IPSSRF} & \texttt{IPSSL1} & \texttt{KOGLM} & \texttt{KORF} & \texttt{KOL1} & \texttt{DeepPINK} & \texttt{SSBoost} \\
\midrule
miR-21 & 190920 & -- & 0.08 & -- & -- & -- & -- & -- & -- \\
miR-155 & 128839 & 0.32 & 0.08 & 0.09 & -- & -- & 0.20 & -- & -- \\
miR-145 & 89265 & -- & 0.24 & -- & -- & -- & -- & -- & -- \\
miR-146a & 57373 & -- & 0.03 & -- & -- & -- & -- & -- & -- \\
miR-214 & 57253 & -- & 0.05 & -- & -- & -- & -- & -- & -- \\
miR-223 & 53770 & 0.24 & 0.03 & 0.09 & -- & -- & -- & -- & -- \\
miR-25 & 43636 & 0.12 & 0.32 & 0.09 & -- & -- & 0.20 & -- & -- \\
miR-22 & 42211 & 0.04 & 0.03 & 0.07 & -- & -- & 0.20 & -- & 0.17 \\
miR-150 & 41195 & 0.04 & 0.03 & 0.07 & -- & -- & 0.20 & -- & 0.17 \\
miR-142 & 39980 & 0.04 & 0.03 & -- & -- & -- & -- & -- & 0.25 \\
miR-335 & 33200 & 0.26 & -- & -- & -- & -- & -- & -- & -- \\
miR-15b & 30988 & 0.24 & -- & 0.20 & -- & -- & 0.25 & -- & -- \\
miR-140 & 29667 & 0.04 & 0.03 & -- & -- & -- & -- & -- & 0.17 \\
miR-152 & 25767 & -- & 0.16 & -- & -- & -- & -- & -- & -- \\
\midrule
$\geq$ 1000 & -- & 14 & 25 & 12 & 0 & 0 & 7 & 0 & 5 \\
$<$ 1000 & -- & 0 & 3 & 1 & 0 & 0 & 1 & 0 & 0 \\
Total & -- & 14 & 28 & 13 & 0 & 0 & 8 & 0 & 5 \\
\bottomrule
\end{tabular}
\caption{\textit{MicroRNAs and tumor purity (ovarian cancer).} MiRNAs are ordered by citation count. A missing $q$-value indicates the miRNA was assigned a $q$-value of less than 0.35 by the corresponding method. The bottom rows report, for each method, the number of selected features with over 1000 citations, under 1000 citations, and the total number selected at the maximum target FDR of 0.35.}
\label{tab:ovarian_mirna_Tumor_purity}
\end{table}

\begin{table}[H]
\centering
\begin{tabular}{lccccccccc}
\toprule
\textbf{Gene} & \textbf{Citations} & \texttt{IPSSGB} & \texttt{IPSSRF} & \texttt{IPSSL1} & \texttt{KOGLM} & \texttt{KORF} & \texttt{KOL1} & \texttt{DeepPINK} & \texttt{SSBoost} \\
\midrule
CD38 & 109641 & 0.13 & 0.14 & -- & -- & -- & -- & -- & -- \\
AKT2 & 97889 & 0.13 & -- & -- & 0.30 & -- & -- & -- & -- \\
ERBB4 & 61018 & -- & -- & -- & -- & -- & 0.45 & -- & -- \\
CCR3 & 25401 & -- & 0.23 & -- & -- & 0.35 & -- & -- & -- \\
CD1C & 24420 & -- & 0.38 & -- & -- & -- & -- & -- & -- \\
WTAP & 22418 & 0.24 & -- & -- & -- & -- & -- & -- & -- \\
SHMT2 & 19844 & -- & -- & -- & 0.30 & -- & -- & -- & -- \\
AAAS & 16226 & -- & -- & -- & 0.30 & -- & 0.45 & -- & -- \\
PAK4 & 14396 & -- & 0.38 & -- & -- & -- & -- & -- & -- \\
SLAMF7 & 12027 & 0.24 & 0.14 & -- & -- & -- & -- & -- & -- \\
\midrule
$\geq$ 200 & -- & 15 & 14 & 8 & 13 & 2 & 7 & 0 & 0 \\
$<$ 200 & -- & 4 & 1 & 3 & 4 & 1 & 0 & 0 & 0 \\
Total & -- & 19 & 15 & 11 & 17 & 3 & 7 & 0 & 0 \\
\bottomrule
\end{tabular}
\caption{\textit{RNA-seq and prognosis (ovarian cancer).} Genes are ordered by citation count. A missing $q$-value indicates the gene was assigned a $q$-value of less than 0.5 by the corresponding method. The bottom rows report, for each method, the number of selected features with over 200 citations, under 200 citations, and the total number selected at the maximum target FDR of 0.5.}
\label{tab:ovarian_rnaseq_status}
\end{table}

\begin{table}[H]
\centering
\begin{tabular}{lccccccccc}
\toprule
\textbf{miRNA} & \textbf{Citations} & \texttt{IPSSGB} & \texttt{IPSSRF} & \texttt{IPSSL1} & \texttt{KOGLM} & \texttt{KORF} & \texttt{KOL1} & \texttt{DeepPINK} & \texttt{SSBoost} \\
\midrule
miR-155 & 94172 & -- & 0.05 & -- & -- & -- & -- & -- & -- \\
miR-10b & 42000 & 0.21 & 0.05 & 0.03 & -- & -- & 0.35 & -- & 0.39 \\
miR-148a & 32559 & -- & 0.13 & -- & -- & -- & -- & -- & -- \\
miR-335 & 22423 & -- & 0.08 & 0.05 & -- & -- & 0.50 & -- & -- \\
miR-15b & 21960 & 0.14 & 0.05 & 0.03 & -- & -- & 0.20 & -- & 0.38 \\
miR-424 & 21680 & -- & 0.37 & -- & -- & -- & -- & -- & -- \\
miR-10a & 20065 & -- & 0.21 & -- & -- & -- & -- & -- & -- \\
miR-224 & 18932 & 0.16 & -- & -- & -- & -- & -- & -- & 0.38 \\
miR-503 & 11894 & 0.14 & 0.06 & -- & -- & -- & -- & -- & 0.38 \\
let-7e & 11348 & 0.25 & 0.06 & -- & -- & -- & -- & -- & 0.38 \\
\midrule
$\geq$ 100 & -- & 15 & 21 & 6 & 0 & 0 & 14 & 0 & 14 \\
$<$ 100 & -- & 5 & 2 & 3 & 0 & 0 & 10 & 0 & 4 \\
Total & -- & 20 & 23 & 9 & 0 & 0 & 24 & 0 & 18 \\
\bottomrule
\end{tabular}
\caption{\textit{MicroRNAs and prognosis (glioma).} MiRNAs are ordered by citation count. A missing $q$-value indicates the miRNA was assigned a $q$-value of less than 0.5 by the corresponding method. The bottom rows report, for each method, the number of selected features with over 100 citations, under 100 citations, and the total number selected at the maximum target FDR of 0.5.}
\label{tab:glioma_mirna_status}
\end{table}

\begin{table}[H]
\centering
\begin{tabular}{lccccccccc}
\toprule
\textbf{Gene} & \textbf{Citations} & \texttt{IPSSGB} & \texttt{IPSSRF} & \texttt{IPSSL1} & \texttt{KOGLM} & \texttt{KORF} & \texttt{KOL1} & \texttt{DeepPINK} & \texttt{SSBoost} \\
\midrule
FOXM1 & 62538 & 0.25 & -- & -- & -- & -- & -- & -- & -- \\
WEE1 & 26648 & 0.10 & 0.06 & -- & -- & 0.44 & -- & -- & -- \\
IGFBP2 & 24482 & -- & 0.08 & -- & -- & -- & -- & -- & -- \\
CX3CL1 & 23044 & -- & -- & -- & -- & -- & 0.34 & -- & -- \\
TIMP1 & 22632 & -- & -- & -- & -- & 0.44 & -- & -- & -- \\
SKI & 19220 & 0.12 & 0.23 & 0.03 & -- & 0.44 & 0.34 & -- & -- \\
CCNB1 & 14856 & -- & 0.10 & -- & -- & -- & -- & -- & -- \\
CDK9 & 14558 & 0.25 & -- & -- & -- & -- & -- & -- & -- \\
TOP2A & 13820 & -- & -- & 0.03 & -- & -- & -- & -- & -- \\
PDPN & 12929 & -- & 0.23 & -- & -- & -- & -- & -- & -- \\
MSN & 11866 & -- & 0.10 & -- & -- & -- & -- & -- & -- \\
ATF2 & 11040 & 0.10 & -- & -- & -- & -- & -- & -- & -- \\
\midrule
$\geq$ 500 & -- & 19 & 18 & 12 & 0 & 22 & 24 & 0 & 0 \\
$<$ 500 & -- & 9 & 6 & 7 & 0 & 15 & 36 & 0 & 0 \\
Total & -- & 28 & 24 & 19 & 0 & 37 & 60 & 0 & 0 \\
\bottomrule
\end{tabular}
\caption{\textit{RNA-seq and prognosis (glioma).} Genes are ordered by citation count. A missing $q$-value indicates the gene was assigned a $q$-value of less than 0.5 by the corresponding method. The bottom rows report, for each method, the number of selected features with over 500 citations, under 500 citations, and the total number selected at the maximum target FDR of 0.5.}
\label{tab:glioma_rnaseq_status}
\end{table}


\subsection{Cross-validation.}\label{sup_sec:cv_study}


As noted in the main text, we also measure feature selection performance by implementing a 20-fold cross-validation (CV) procedure, described as follows. In each of the 20 CV steps, one group of patients is set aside (the test set), and a set of features is selected by each method using the data in the remaining groups (the training set). Next, for each method, we construct three predictive models---a linear model, a random forest model, and a gradient boosting model---using only the features selected by that method on the training data. Each model is then used to predict responses from the test set, and the smallest of the three prediction errors is recorded (we use mean squared error for regression and $1 - \mathrm{accuracy}$ for classification). All three models are implemented to ensure that no method has an inherent advantage over another. For example, the features selected by \texttt{IPSSL1} may be better suited to minimizing error in a linear model than those selected by \texttt{IPSSGB}, while those selected by \texttt{IPSSGB} may be better suited to minimizing error in a gradient boosting model than the ones selected by \texttt{IPSSL1}. The linear and random forest predictive models are implemented with scikit-learn \citep{sklearn} and gradient boosting with \texttt{XGBoost} \citep{xgboost}, always with default parameters. For continuous responses, in each CV step we subtract the mean of the training responses from all responses, training and test, and scale all responses by the empirical standard deviation of the training responses.

CV study results are shown in Figures~\ref{fig:cv_study_ovarian_mirna_Tumor_purity}--\ref{fig:cv_study_glioma_rnaseq_FOXM1}. Each plot contains two subplots. The left subplots show the prediction error associated to the features selected by each method at the given target FDR. In all studies, we find that the three IPSS methods select features at much lower target FDRs than all of the model-X knockoffs methods. The right subplots show the prediction error as a function of the number of features selected by each method. Curves for each method in these plots are obtained by varying the target FDR between 0 and 0.5. \texttt{Boruta} does not have FDR control parameters and is therefore represented by a single point in these plots. In each plot, the dashed black line shows the average error when using all features in the dataset to predict the response variable. \texttt{DeepPINK} rarely selects any features and is therefore represented by a single point for better visibility.

\ifthenelse{\boolean{showfigures}}{
\begin{figure}[ht]
\makebox[\textwidth][c]{%
	{\includegraphics[height=.25\textheight, width=\textwidth]{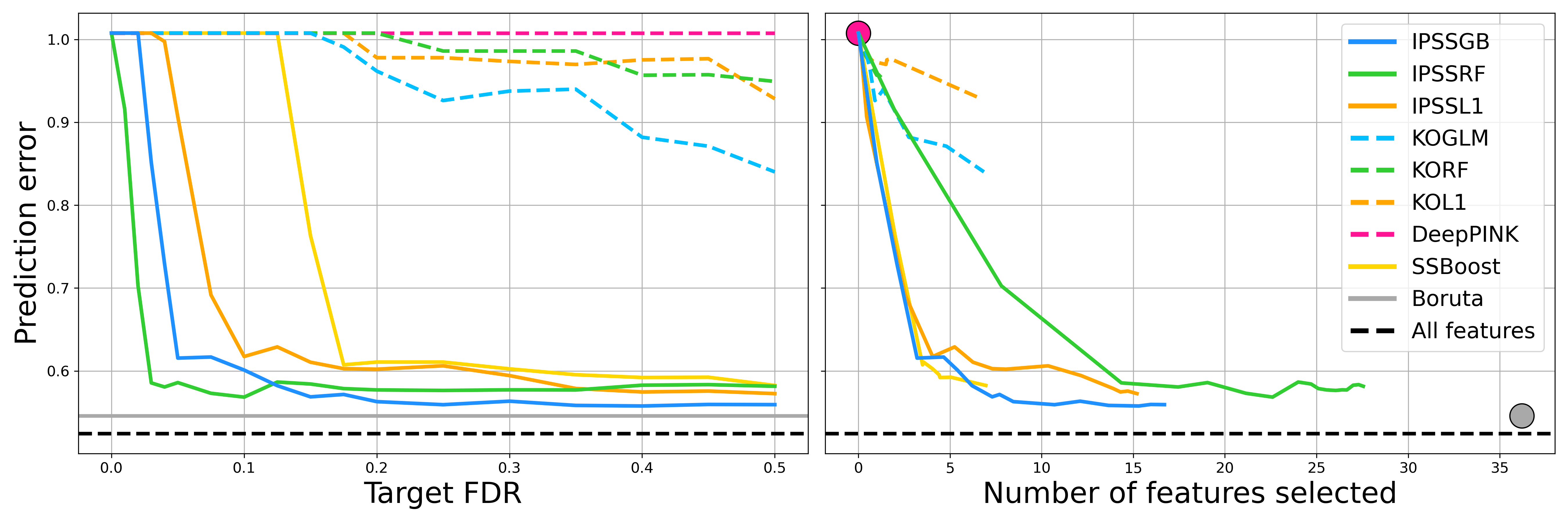}}%
}
\caption{\textit{MiRNAs and tumor purity (ovarian cancer).}}
\label{fig:cv_study_ovarian_mirna_Tumor_purity}
\end{figure}
}{}

\ifthenelse{\boolean{showfigures}}{
\begin{figure}[ht]
\makebox[\textwidth][c]{%
	{\includegraphics[height=.25\textheight, width=\textwidth]{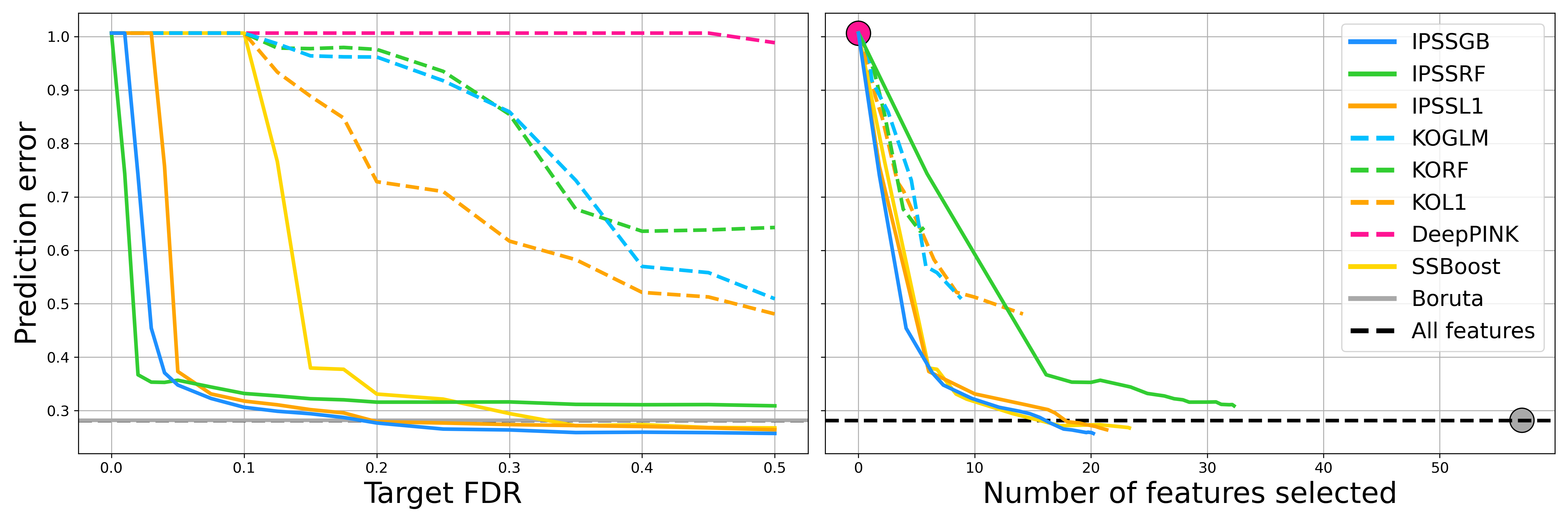}}%
}
\caption{\textit{MiRNAs and miR-150 (ovarian cancer).}}
\label{fig:cv_study_ovarian_mirna_hsa-mir-150}
\end{figure}
}{}

\ifthenelse{\boolean{showfigures}}{
\begin{figure}[ht]
\makebox[\textwidth][c]{%
	{\includegraphics[height=.25\textheight, width=\textwidth]{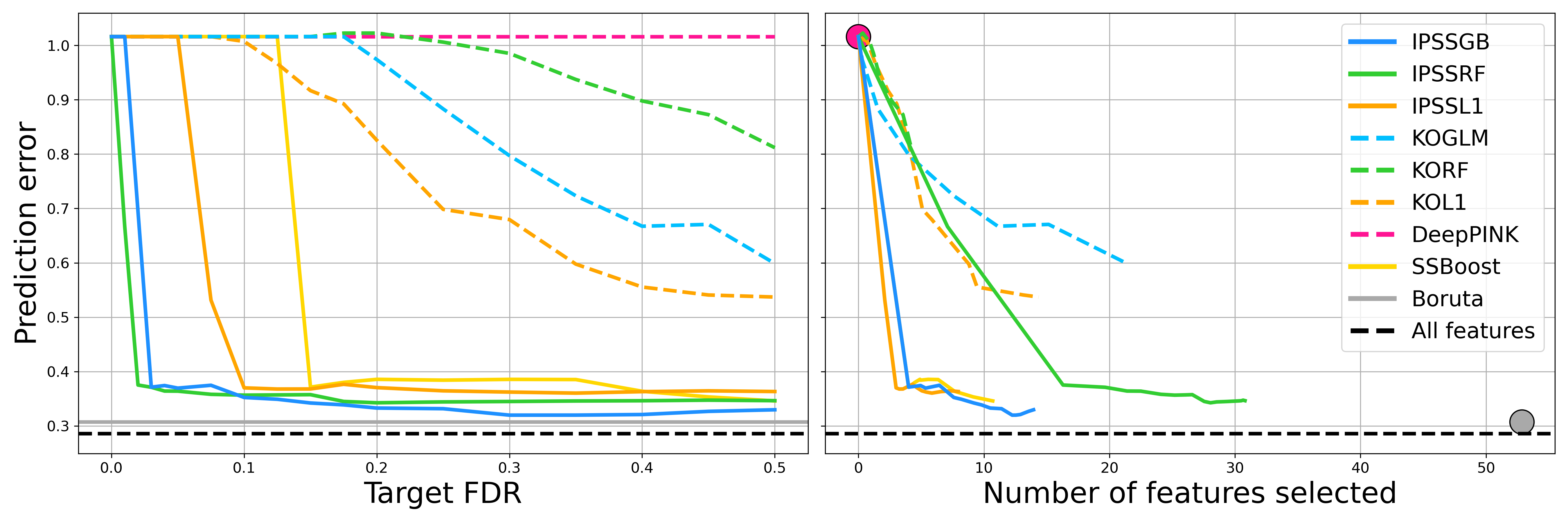}}%
}
\caption{\textit{Genes and AKT2 (ovarian cancer).}}
\label{fig:cv_study_ovarian_rnaseq_AKT2}
\end{figure}
}{}

\ifthenelse{\boolean{showfigures}}{
\begin{figure}[ht]
\makebox[\textwidth][c]{%
	{\includegraphics[height=.25\textheight, width=\textwidth]{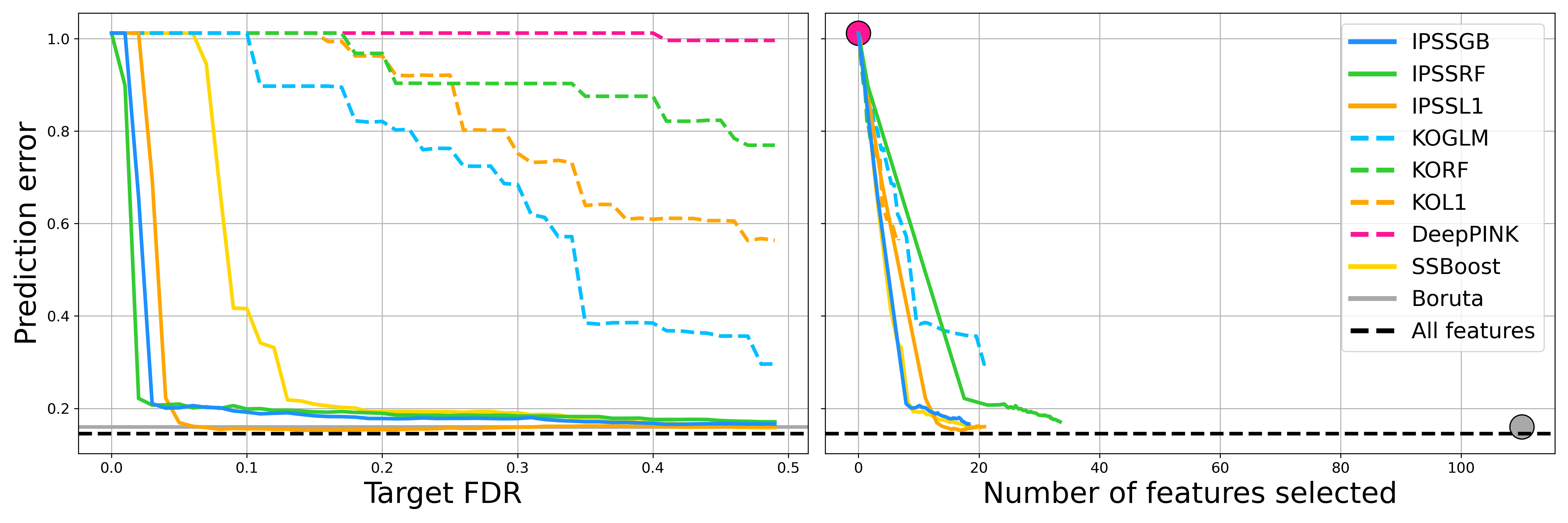}}%
}
\caption{\textit{MiRNAs and miR-155 (glioma).}} 
\label{fig:cv_study_glioma_mirna_hsa-mir-155}
\end{figure}
}{}

\ifthenelse{\boolean{showfigures}}{
\begin{figure}[ht]
\makebox[\textwidth][c]{%
	{\includegraphics[height=.25\textheight, width=\textwidth]{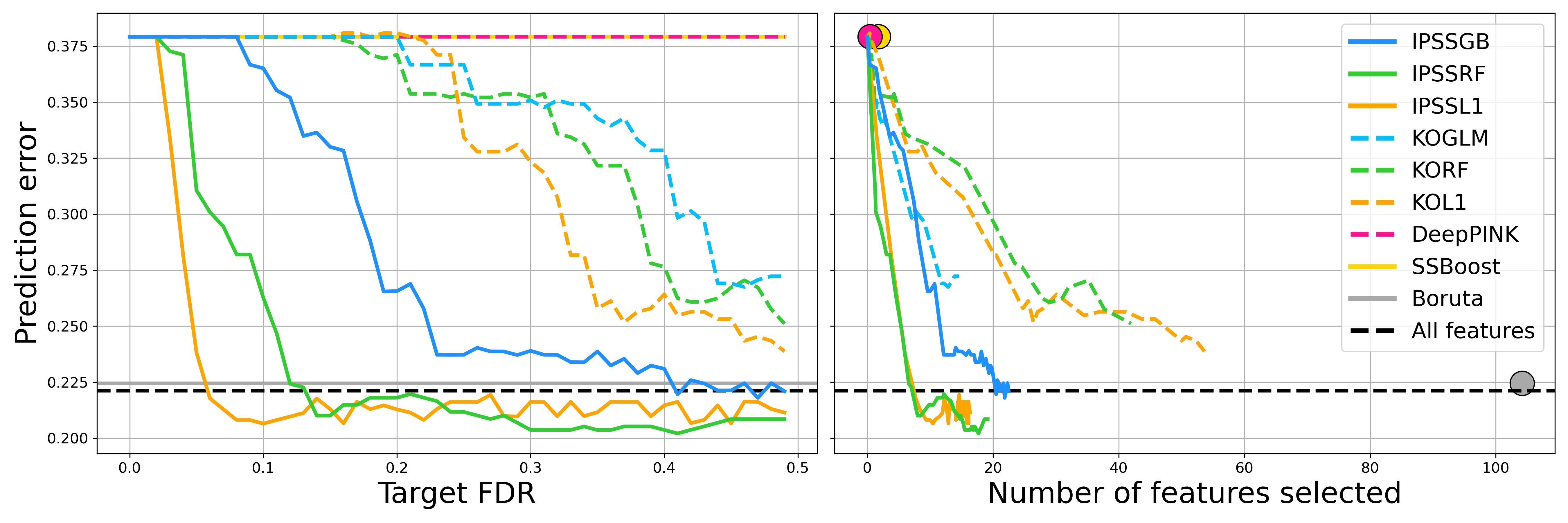}}%
}
\caption{\textit{Genes and prognosis (glioma).}}
\label{fig:cv_study_glioma_rnaseq_status}
\end{figure}
}{}

\ifthenelse{\boolean{showfigures}}{
\begin{figure}[ht]
\makebox[\textwidth][c]{%
	{\includegraphics[height=.25\textheight, width=\textwidth]{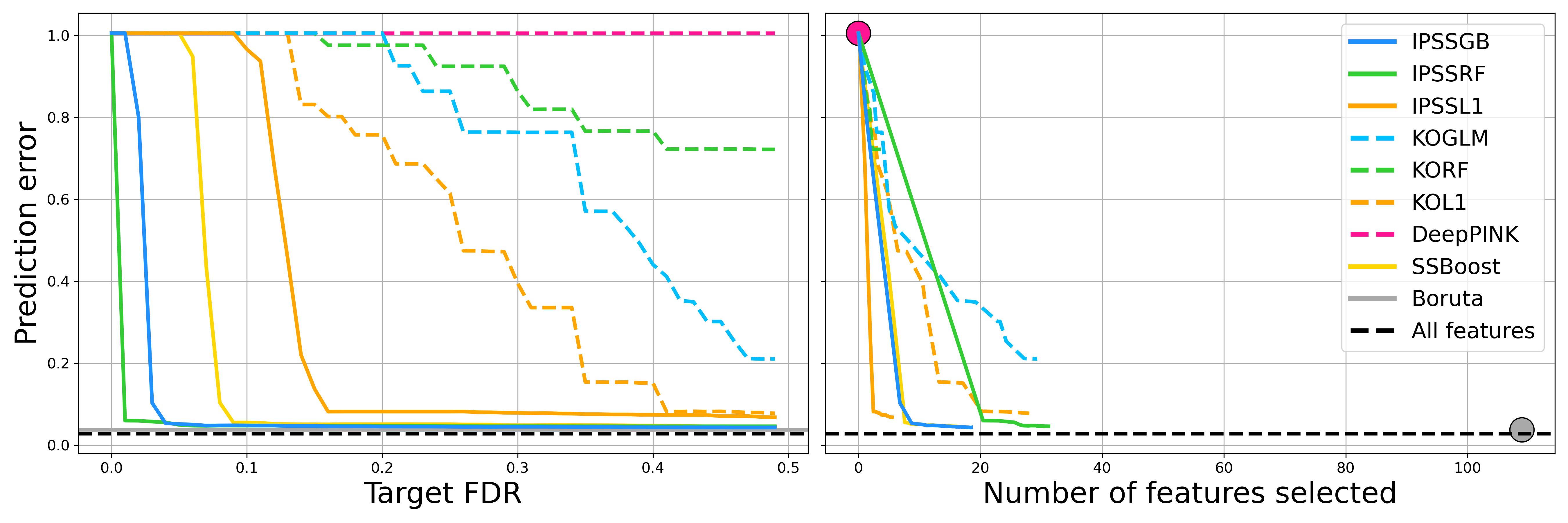}}%
}
\caption{\textit{Genes and FOXM1 (glioma).}}
\label{fig:cv_study_glioma_rnaseq_FOXM1}
\end{figure}
}{}

\clearpage


\section{Sensitivity analyses}\label{sup_sec:sensitivity}


Figures~\ref{fig:sensitivity_cutoff_reg}--\ref{fig:sensitivity_f_class} show the sensitivity of \texttt{IPSSGB} to the IPSS parameters discussed in \cref{sec:methods,sup_sec:parameters}. Data are simulated according to the ovarian cancer RNA-seq simulation design described in \cref{sec:rnaseq_simulations,sup_sec:simulation_details} for both regression and classification. The results for \texttt{IPSSRF} were similar and are therefore omitted.

\ifthenelse{\boolean{showfigures}}{
\begin{figure}[ht]
\makebox[\textwidth][c]{%
	{\includegraphics[height=.325\textheight, width=\textwidth]{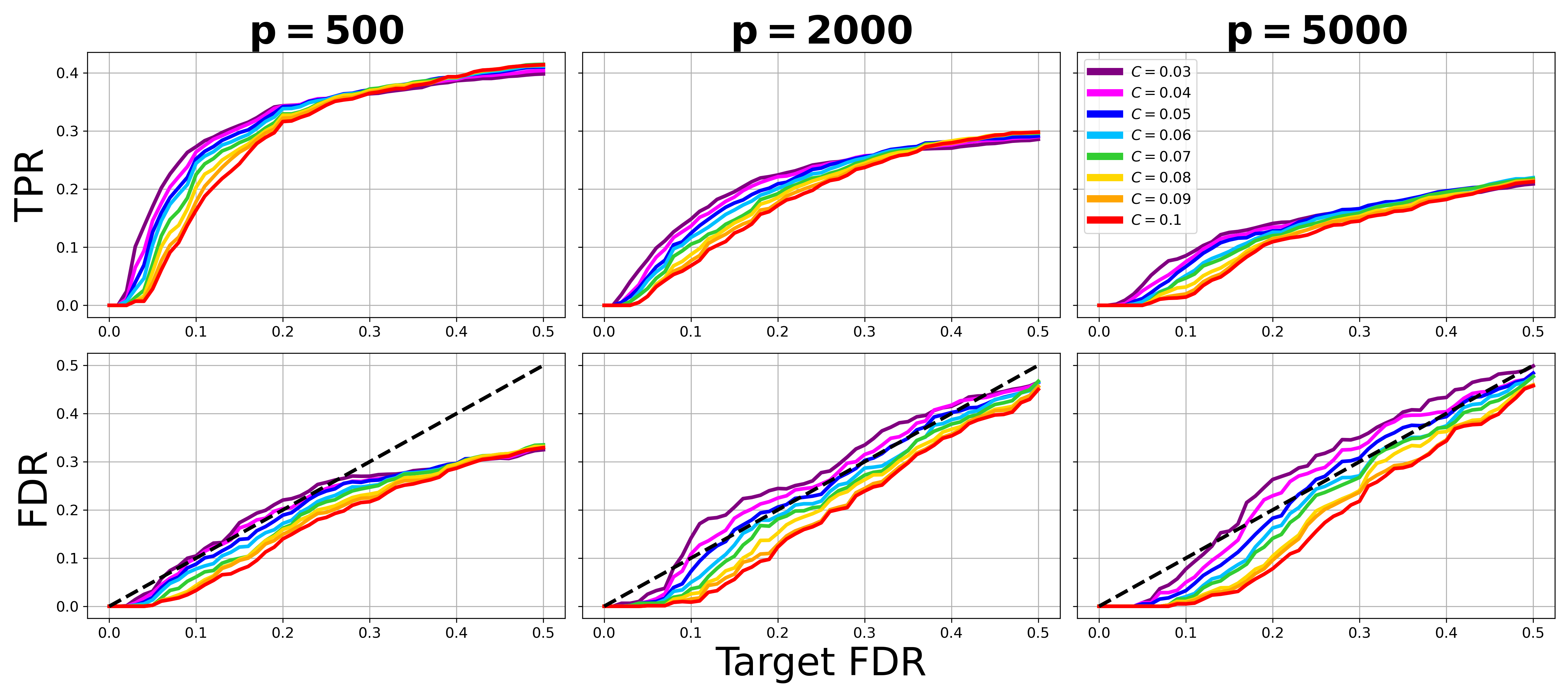}}%
}
\caption{\textit{Different choices of $C$ (regression).}}
\label{fig:sensitivity_cutoff_reg}
\end{figure}
}{}

\ifthenelse{\boolean{showfigures}}{
\begin{figure}[ht]
\makebox[\textwidth][c]{%
	{\includegraphics[height=.325\textheight, width=\textwidth]{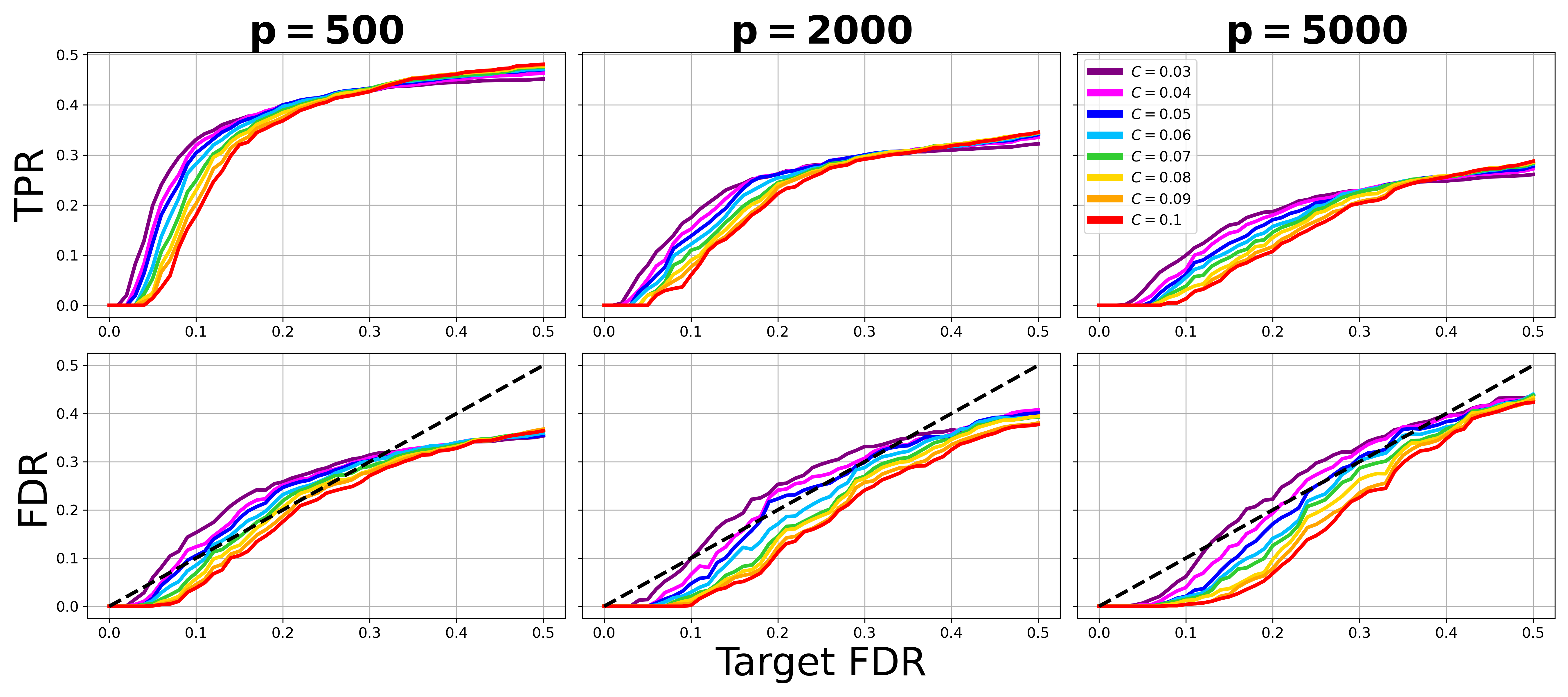}}%
}
\caption{\textit{Different choices of $C$ (classification).}}
\label{fig:sensitivity_cutoff_class}
\end{figure}
}{}

\ifthenelse{\boolean{showfigures}}{
\begin{figure}[ht]
\makebox[\textwidth][c]{%
	{\includegraphics[height=.325\textheight, width=\textwidth]{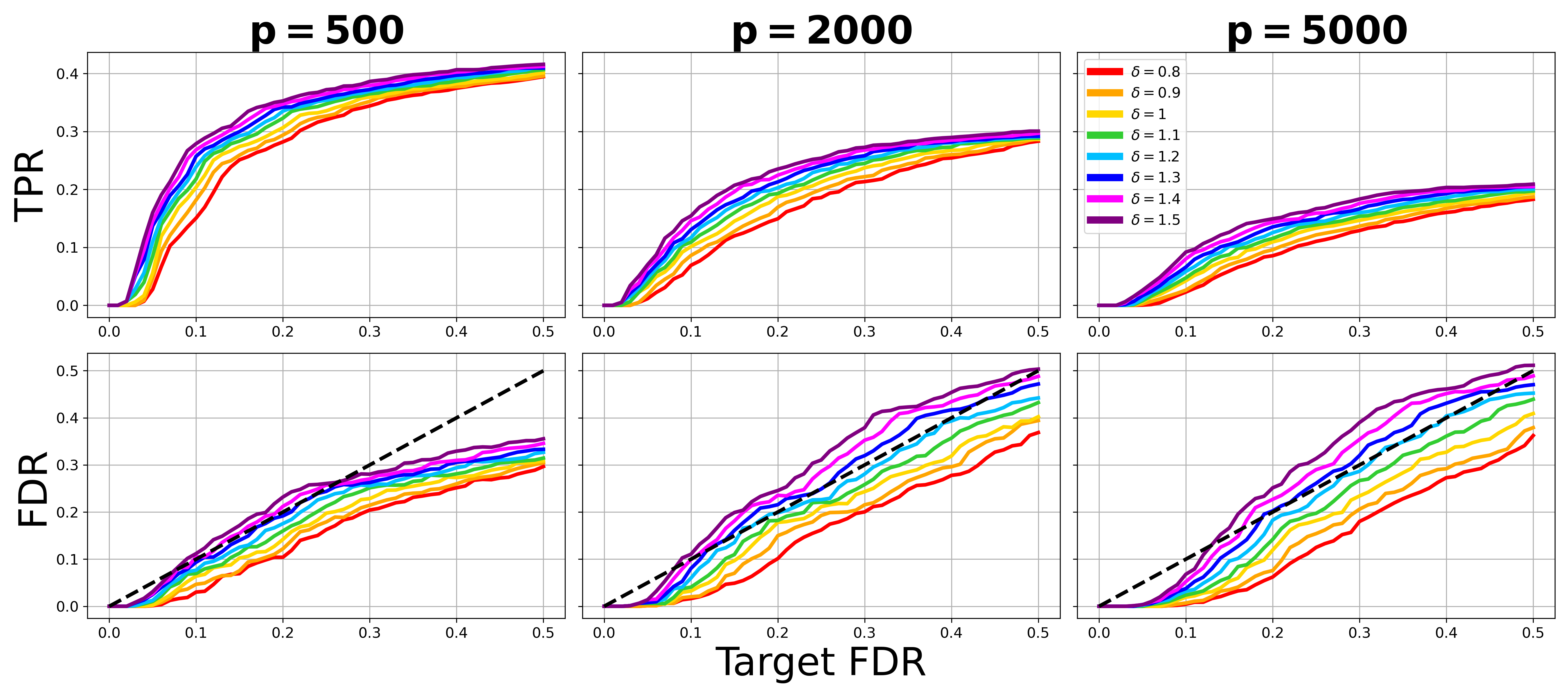}}%
}
\caption{\textit{Different choices of $\delta$ (regression).}}
\label{fig:sensitivity_delta_reg}
\end{figure}
}{}

\ifthenelse{\boolean{showfigures}}{
\begin{figure}[ht]
\makebox[\textwidth][c]{%
	{\includegraphics[height=.325\textheight, width=\textwidth]{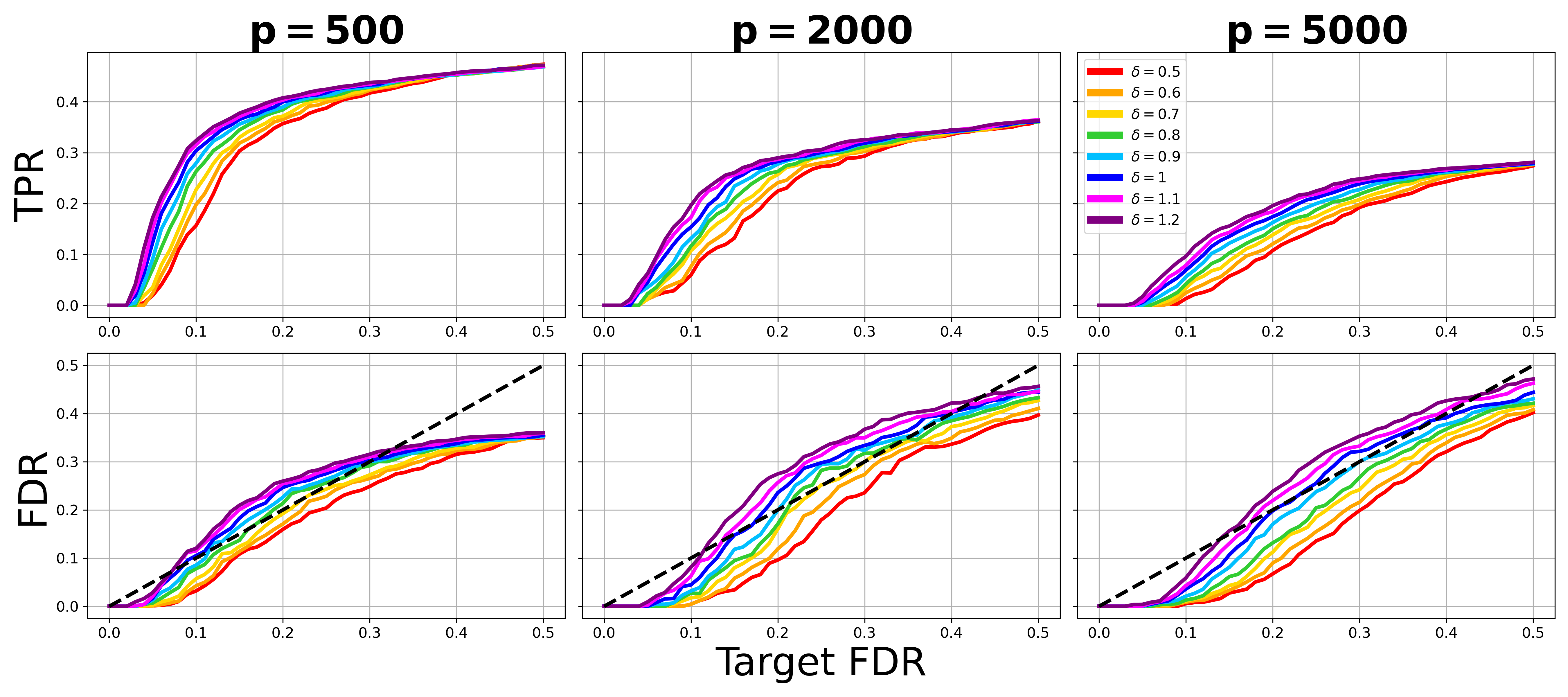}}%
}
\caption{\textit{Different choices of $\delta$ (classification).}}
\label{fig:sensitivity_delta_class}
\end{figure}
}{}

\ifthenelse{\boolean{showfigures}}{
\begin{figure}[ht]
\makebox[\textwidth][c]{%
	{\includegraphics[height=.325\textheight, width=\textwidth]{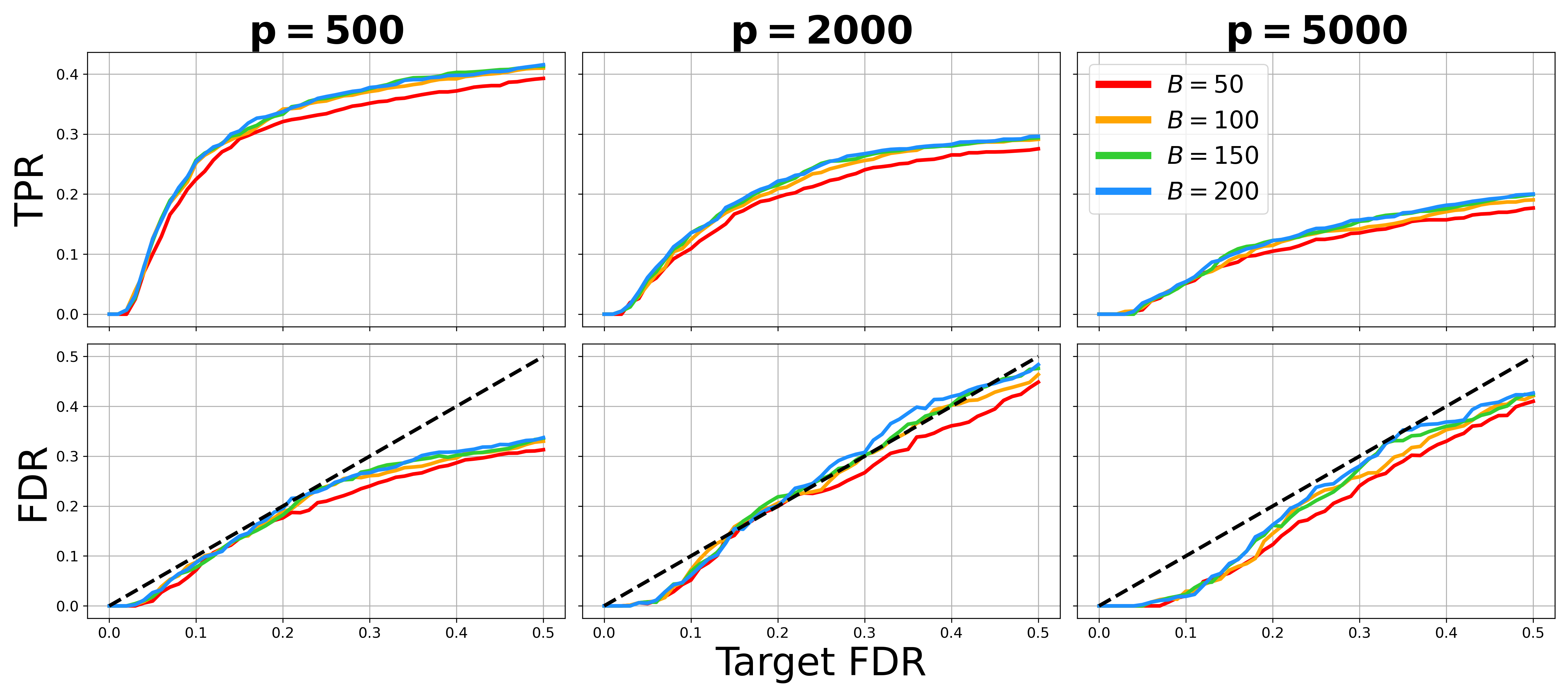}}%
}
\caption{\textit{Different choices of $B$ (regression).}}
\label{fig:sensitivity_B_reg}
\end{figure}
}{}

\ifthenelse{\boolean{showfigures}}{
\begin{figure}[ht]
\makebox[\textwidth][c]{%
	{\includegraphics[height=.325\textheight, width=\textwidth]{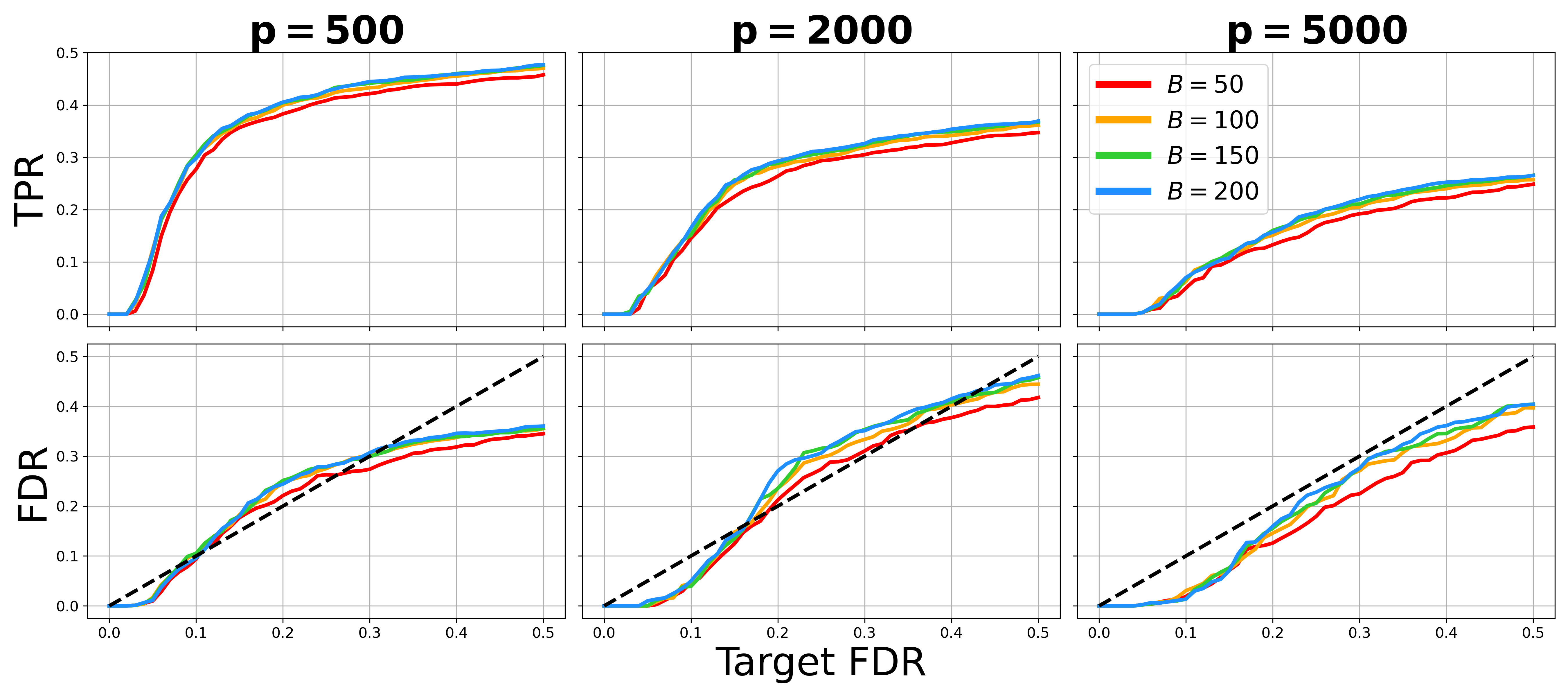}}%
}
\caption{\textit{Different choices of $B$ (classification).}}
\label{fig:sensitivity_B_class}
\end{figure}
}{}

\ifthenelse{\boolean{showfigures}}{
\begin{figure}[ht]
\makebox[\textwidth][c]{%
	{\includegraphics[height=.325\textheight, width=\textwidth]{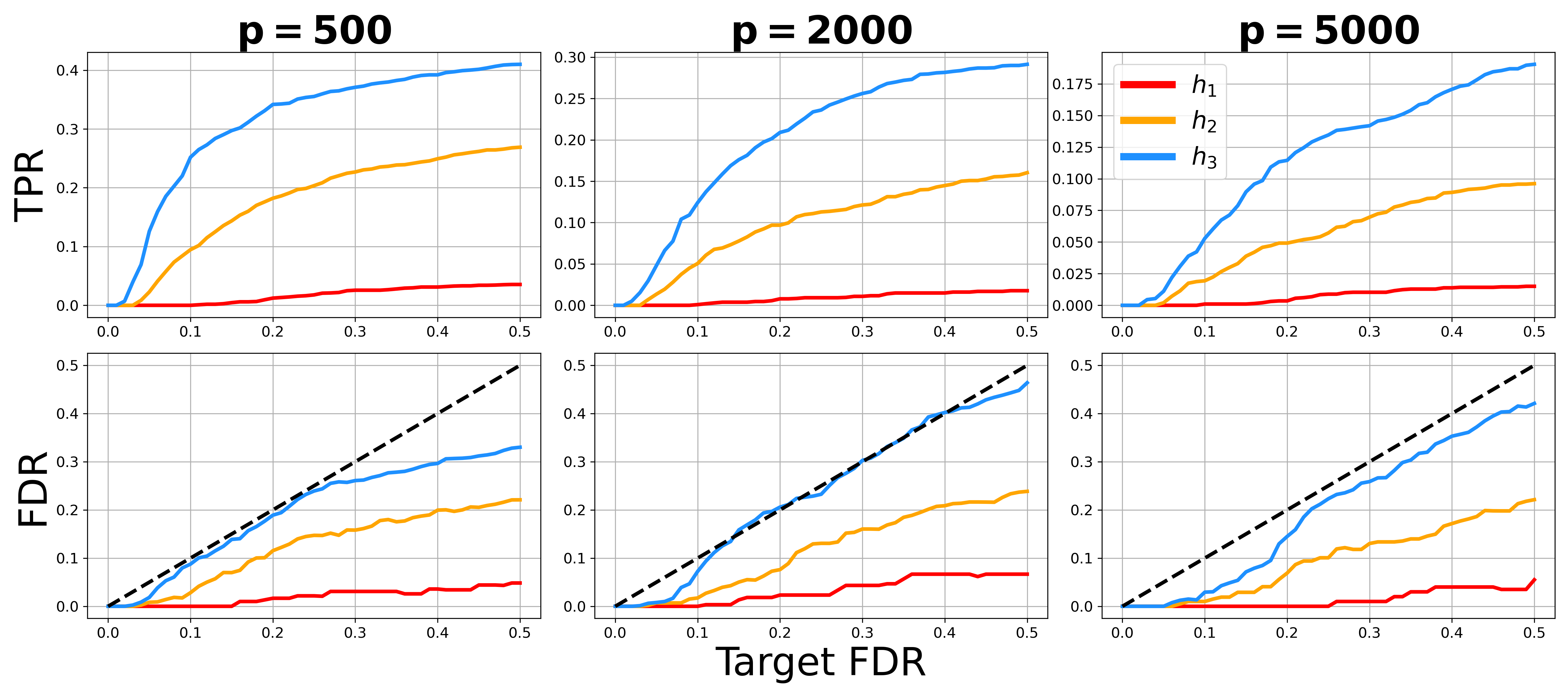}}%
}
\caption{\textit{Different choices of function (regression).}}
\label{fig:sensitivity_f_reg}
\end{figure}
}{}

\ifthenelse{\boolean{showfigures}}{
\begin{figure}[ht]
\makebox[\textwidth][c]{%
	{\includegraphics[height=.325\textheight, width=\textwidth]{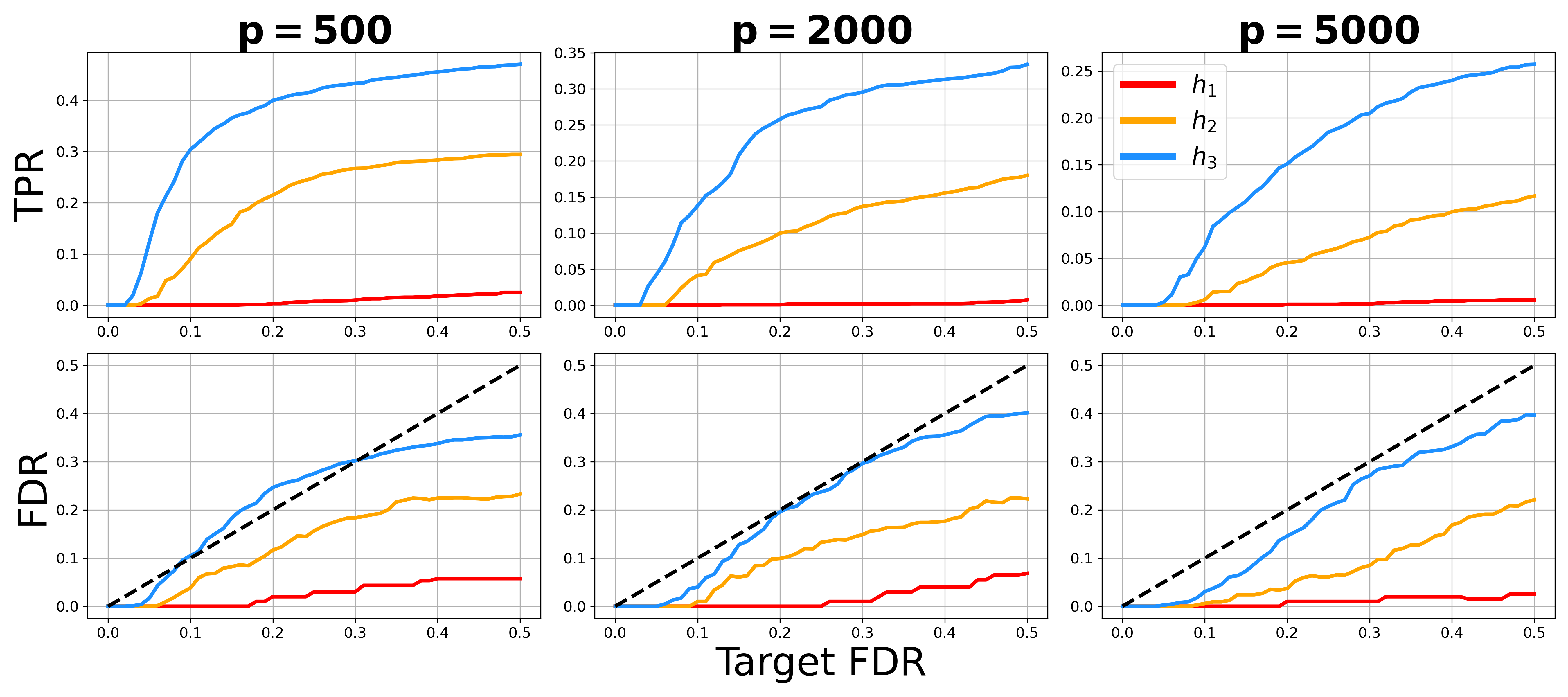}}%
}
\caption{\textit{Different choices of function (classification).}}
\label{fig:sensitivity_f_class}
\end{figure}
}{}

\end{document}